\documentclass[a4paper,fleqn]{cas-dc}
\usepackage{amsmath}
\AtBeginDocument{\color{black}}
\usepackage[numbers]{natbib}
\usepackage{algorithm}
\usepackage{algpseudocode}

\def\tsc#1{\csdef{#1}{\textsc{\lowercase{#1}}\xspace}}
\tsc{WGM}
\tsc{QE}
\tsc{EP}
\tsc{PMS}
\tsc{BEC}
\tsc{DE}
\shorttitle{Balanced Diffusion-Guided Fusion for Multimodal Remote Sensing Classification}
\shortauthors{Liu et al.}

\begin{document}
\let\WriteBookmarks\relax
\def\floatpagepagefraction{1}
\def\textpagefraction{.001}

\title [mode = title]{Balanced Diffusion-Guided Fusion for Multimodal Remote Sensing Classification}

\author[1]{Hao Liu}
\credit{Conceptualization of this study, Methodology, Software}
\author[1]{Yongjie Zheng}
\author[2]{Yuhan Kang}
\author[3,4]{Mingyang Zhang}
\credit{Data curation, Writing - Original draft preparation}
\author[3,4,5]{Maoguo Gong}
\author[1]{Lorenzo Bruzzone\cormark[1]}

\affiliation[1]{organization={Department of Information Engineering and Computer Science, University of Trento},
	city={Trento},
	postcode={38123}, 
	country={Italy}}
\affiliation[2]{organization={College of Electrical and Information Engineering, Hunan University}, 
	city={Changsha},
	postcode={410082}, 
	country={China}}
\affiliation[3]{organization={Key Laboratory of Collaborative Intelligent Systems of Ministry of Education, Xidian University},
	city={Xi’an},
	postcode={710071}, 
	country={China}}
\affiliation[4]{organization={School of Electronic Engineering, Xidian University},
	city={Xi’an},
	postcode={710071}, 
country={China}}
\affiliation[5]{organization={Academy of Artificial Intelligence, Inner Mongolia Normal University},
	city={Hohhot},
	postcode={010028}, 
	country={China}}

\cortext[cor1]{Corresponding author}

\nonumnote{Email addresses: hao.liu@unitn.it; yongjie.zheng@unitn.it; kyh433@hnu.edu.cn; omegazhangmy@gmail.com; gong@ieee.org; lorenzo.bruzzone@unitn.it.
  }

\begin{abstract}
Deep learning-based techniques for the analysis of multimodal remote sensing data have become popular due to their ability to effectively integrate spatial, spectral, and structural information from different sensors. Recently, denoising diffusion probabilistic models (DDPMs) have attracted attention in the remote sensing community due to their powerful ability to capture robust and complex data distributions. However, pre-training multimodal DDPMs may result in modality imbalance, and effectively leveraging diffusion features to guide complementary diversity feature extraction remains an open question. To address these issues, this paper proposes a balanced diffusion-guided fusion framework that leverages multimodal diffusion features to guide a group network for land-cover classification. Specifically, we propose an adaptive modality masking strategy to encourage the DDPMs to obtain a modality-balanced rather than spectral image-dominated data distribution. Subsequently, these diffusion features hierarchically guide the extraction of local, sequential, and global features by integrating feature fusion, group channel attention, and cross-attention mechanisms. Finally, a mutual learning strategy is developed to enhance inter-branch collaboration by aligning the probability entropy and similarity of diverse features. Extensive experiments on four multimodal remote sensing datasets demonstrate that the proposed method achieves superior classification performance. The code is available at https://github.com/HaoLiu-XDU/BDGF.
\end{abstract}

%\begin{graphicalabstract}
%\includegraphics{figs/cas-grabs.pdf}
%\end{graphicalabstract}

%\begin{highlights}
%\item Research highlights item 1
%\item Research highlights item 2
%\item Research highlights item 3
%\end{highlights}


\maketitle

\section{Introduction}
\label{sec1}	

Multimodal remote sensing data typically comprises images from various sensors, including hyperspectral images (HSI), synthetic aperture radar (SAR), and light detection and ranging. By fusing complementary spatial, spectral, and structural information, multimodal data facilitates an accurate classification \cite{10778974}, thereby supporting applications in environmental monitoring \cite{he2017environmental,9740200}, agricultural planning \cite{zheng2024new}, and resource exploration \cite{zhang2010multi}.

In recent years, deep learning methods have revolutionized multimodal remote sensing image classification. Regarding feature extraction and fusion strategy, network architectures are primarily based on convolutional neural networks (CNNs), Transformers, and Mamba models. CNNs serve as the foundation of many multimodal frameworks due to their ability to capture local spatial patterns through convolutional filters \cite{7115053,10517881}. Transformers can capture global context through self-attention mechanisms \cite{gao2022fusion}. Moreover, the Mamba model \cite{gu2023mamba} has gained popularity for capturing long-range dependencies with improved computational efficiency \cite{10856240,ahmad2024comprehensive}. In terms of multimodal fusion strategies, Transformer-based methods typically employ global cross-attention mechanisms at the token level, while Mamba-based approaches optimize Mamba blocks to simultaneously perform multimodal fusion and model long-range dependencies. Adaptively merging these local, global, and sequence-aware representations can leverage their strengths and compensate for their weaknesses \cite{chen2021remote,lu2023coupled,li2023mixing}.

However, these existing networks are primarily optimized to extract discriminative features for specific downstream tasks and they often discard broader contextual information. This prevents models from capturing the complex, high-dimensional, and redundant distributions characteristic of remote sensing images. In addition, these approaches still struggle to model complementary and diverse features \cite{10679212,10750894} and can be brittle in the presence of realistic sensor noise, atmospheric effects, or occlusions \cite{ahmad2024comprehensive}.

Diffusion-based two-stage methods have emerged as a promising method for remote sensing image interpretation \cite{10684806}. By iteratively refining noisy inputs, denoising diffusion probabilistic models (DDPMs) learn robust and noise-reduced representations that capture complex data distributions \cite{ho2020denoising,mukhopadhyay2023diffusion}. For example, Chen et al. \cite{10179942} and Chen et al. \cite{10234379} both employed a diffusion model to extract high-dimensional and redundant distribution from HSI. In \cite{sigger2024unveiling} and \cite{10542168}, the multi-step iterative denoising process was used for fusing multi-time-step features for HSI classification. In multimodal scenarios, Zhang et al. \cite{10716525} only integrated HSI diffusion features instead of multimodal information into the downstream network, which is essentially hyperspectral feature extraction and classification. Typically, all these methods assume that the remote sensing images contain Gaussian noise, which makes it difficult to model noise-robust features for multimodal images, especially in SAR images affected by speckle noise \cite{yang2026self,cao2026global}. In addition, while merging multimodal images into a single pre-training network can reduce computational complexity \cite{10314566}, this approach inadvertently introduces a severe modality imbalance. Specifically, since spectral images inherently carry richer spectral information than SAR data, the HSI modality frequently tends to dominate the optimization process \cite{7169562,Wang_2020_CVPR, 10694738}.

The limitations can be summarized as follows:
\begin{enumerate}
	\item DDPMs capture complex remote sensing distributions through a denoising process. However, uniformly added Gaussian noise does not conform to the imaging mechanisms of all remote sensing images. In addition, few studies have employed global and robust diffusion information to guide the extraction of various characteristics.
	
	\item Concatenating the image's channel dimension into a single network can reduce the computational complexity of the pre-trained model \cite{10314566}. However, this approach introduces a severe modality imbalance problem, as spectral images typically contain more data than SAR images, causing HSIs to dominate the optimization process.
\end{enumerate}
\begin{figure}[t]
	\centering
	\includegraphics[width=\linewidth]{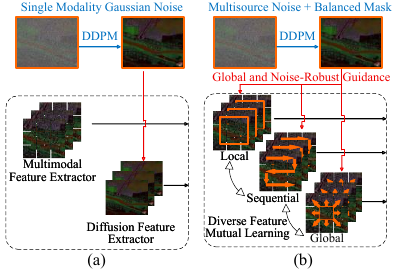}
	\caption
	{Comparison of workflows. (a) Previous methods process diffusion and multimodal features separately and then combine them for joint classification. (b) In this work, we exploit global and noise-robust diffusion information to guide the mutual learning of local, sequence-level, and global features.}
	\label{Motivation}
\end{figure}

Fig. \ref{Motivation} compares the workflows and illustrates the motivation behind our approach. To address these challenges, this paper introduces a balanced diffusion-guided fusion (BDGF) method that employs a pre-training DDPM to guide a group network for multimodal land-cover classification. Initially, an adaptive modality masking strategy is proposed to pre-train the diffusion model and obtain modality-balanced multimodal diffusion features. Subsequently, these diffusion features guide to capture local, global, and sequential information through feature fusion, group channel attention, and cross-attention fusion. Finally, a mutual learning approach dynamically enhances collaboration among the branches and achieves classification based on the probability entropy and feature similarity of each sub-network. The main contributions of this paper are as follows:

\begin{enumerate}
	\item We propose an adaptive modality masking strategy for DDPMs that encourages the model to focus on all information source images by progressively masking dominant images, thereby obtaining diffusion features that reflect the complex intrinsic distribution of multimodal data.
	
	\item We introduce a diffusion-guided fusion strategy that leverages multimodal diffusion information to hierarchically guide the fusion of diverse features based on the global characteristics of different branches.
	
	\item We present a mutual learning strategy that dynamically promotes feature complementarity among networks through predicted probability entropy and pairwise similarity.
	
\end{enumerate}

The remainder of this paper is organized as follows. Section II provides background on the BDGF methodology. Section III describes the proposed method in detail. Section IV validates the method on four real remote sensing datasets. Finally, Section V draws the conclusions of this paper.

\section{Related Work}
This section first reviews the background and advanced methods based on diffusion models in remote sensing, followed by a discussion of techniques for multimodal remote sensing feature fusion.

\subsection{Denoising Diffusion Probabilistic Model}
The DDPM \cite{ho2020denoising} is a widely adopted deep generative model \cite{mukhopadhyay2023diffusion}. Pre-training a DDPM involves both a forward diffusion process and a reverse diffusion process. Given an image $\boldsymbol{x_0}$, the noisy image $\boldsymbol{x_t}$ at time $t$ can be obtained through a Markov chain:
\begin{align}
	\begin{cases}
		\boldsymbol{x}_t = \sqrt{\bar{\alpha}_t} \boldsymbol{x}_0 + \sqrt{1 - \bar{\alpha}_t} \, \epsilon,\\
		\bar{\alpha}_t = \prod_{i=1}^{t} \alpha_i,
		\label{2-1-1}
	\end{cases}
\end{align}
where $\epsilon$ represents the noise and $\alpha_t$ denotes the noise scaling factor. The transfer probability $q(\boldsymbol{x_t}|\boldsymbol{x_{t-1}})$ is defined as:
\begin{align}
	q(\boldsymbol{x}_t | \boldsymbol{x}_{t-1}) = \mathcal{N}(\boldsymbol{x}_t; \sqrt{\alpha_t} \boldsymbol{x}_{t-1}, (1 - \alpha_t)\mathbf{I}),
	\label{2-1-2}
\end{align}
where $\mathbf{I}$ is the identity matrix. Here, $\mu_t=\sqrt{\alpha_t} \boldsymbol{x_{t-1}}$ and $\sigma_t^2 = (1-\alpha_t)\mathbf{I}$ are the mean value and the variance, respectively.

In the reverse process, the model predicts the clean data $\boldsymbol{x_{t-1}}$ given $\boldsymbol{x_t}$. The conditional probability is expressed as:
\begin{align}
	p_\theta(\boldsymbol{x}_{t-1} | \boldsymbol{x}_t) = \mathcal{N}(\boldsymbol{x}_{t-1}; \mu_\theta(\boldsymbol{x}_t, t), \sigma^2_\theta(\boldsymbol{x}_t, t)),
	\label{2-1-3}
\end{align}
where $\mu_\theta$ and $\sigma^2_\theta$ are the predicted mean and variance, parameterized by the model. The predicted mean $\mu_\theta(\boldsymbol{x}_t, t)$ is typically defined as:
\begin{align}
	\mu_\theta(\boldsymbol{x}_t, t) = \frac{1}{\sqrt{\alpha_t}} \left(\boldsymbol{x}_t - \frac{1 - \alpha_t}{\sqrt{1 - \bar{\alpha}_t}} \epsilon_\theta(\boldsymbol{x}_t, t)\right),
	\label{2-1-4}
\end{align}
where $\epsilon_\theta(\boldsymbol{x}_t, t)$ represents the noise predicted by the model.

DDPMs can recover the target distribution through iterative denoising, thereby effectively promoting remote sensing image interpretation tasks \cite{10684806}. Many studies utilize DDPMs to capture the data distribution of complex images and integrate the extracted features at one \cite{10234379} or multiple time steps \cite{10542168} into various tasks during training, such as classification \cite{10179942,sigger2024unveiling}, change detection \cite{jia2024siamese,zhang2023diffucd}, image matching \cite{11024126}, and object detection \cite{chen2023diffusiondet}. In the content of multimodal data analysis, Zhang et al. \cite{10716525} fused the HSI image features extracted by DDPMs into a downstream multi-branch network. Du et al. \cite{10733944} integrated diffusion feature into mamba structure for semantic segmentation.

However, these methods rarely leverage the global guidance provided by diffusion features. Besides, solely extracting diffusion features from HSIs does not satisfy the requirements for multimodal classification. To address these issues, this paper proposes an improved DDPM to obtain modality-balanced multimodal diffusion features.

\begin{figure*}[thpb]
	\centering
	\includegraphics[width=\linewidth]{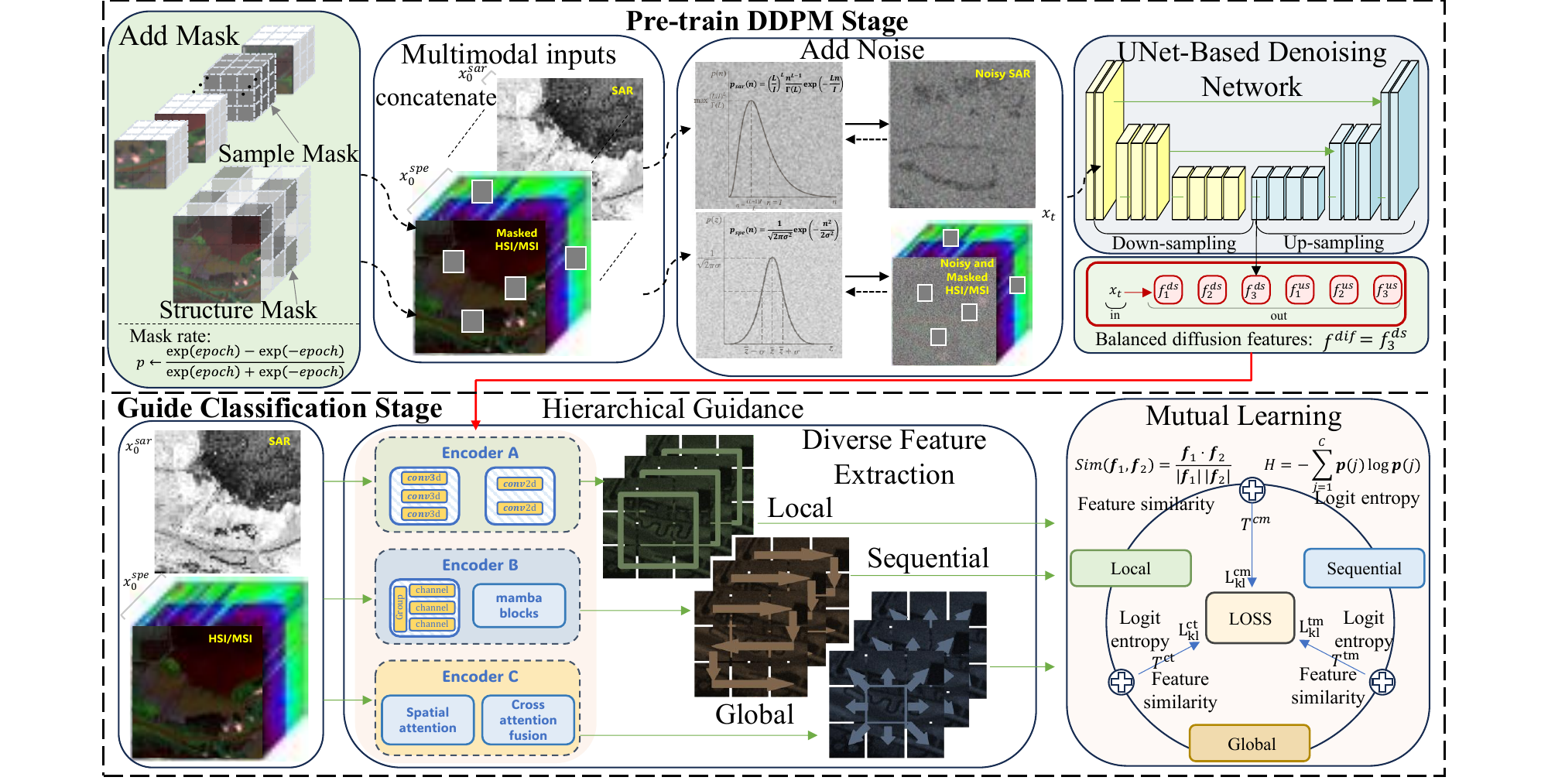}
	\caption
	{Illustration of the proposed BDGF framework.}
	\label{Framwork}
\end{figure*}
\subsection{Multimodal Classification}

The heterogeneity of multimodal remote sensing data prevents a single model from effectively capturing both local features and global dependencies \cite{li2022deep,chen2021remote}. To address this limitation, numerous hybrid algorithms have been developed to leverage the strengths of different architectures, such as CNNs' local inductive bias \cite{9174822,9598903}, transformers' global attention mechanisms \cite{10153685}, and Mamba’s efficient long-sequence modeling \cite{10856240}. In addition, multi-branch architectures enable independent encoding and interactive fusion of heterogeneous modalities, significantly enhancing the model’s ability to capture cross-modal semantic associations.

Several studies have explored the above-mentioned architectures. Gao et al. \cite{gao2022fusion} combined CNNs for local feature extraction with transformers for global modeling. Yang et al. \cite{yang2025d3gnn} used topological structure and convolution network for multimodal remote sensing image classification. Xue et al. \cite{9755059} embedded convolutional operations into spatial and spectral hierarchical transformers to capture both global and local features. Tu et al. \cite{tu2024ncglf2} proposed a fusion strategy based on multi-scale information and a dual-branch structure to integrate global and local representations. Zhang et al. \cite{10738515} introduced Cross-SSM, which extracts multimodal state information by integrating CNNs and Mamba. Liao et al. \cite{10679212} mitigated feature diversity challenges by developing a multimodal classification network incorporating multiple architectural structures.

These methods extract diverse global and local features through hybrid or parallel structures, improving feature extraction and classification. In this paper, we propose integrating robust and complex diffusion features as guiding knowledge into a group network and further enhance the collaboration of diverse features through mutual learning.
\section{Methodology}
The proposed BDGF is designed to use diffusion distribution to guide the complementarity of diverse features. Fig. \ref{Framwork} illustrates the overall BDGF framework. In the pre-training phase, the denoising network generates modality-balanced diffusion features using an adaptive modality masking strategy. During training, these features hierarchically guide the extraction and fusion of information within the group network. Finally, the mutual learning module enhances the alignment of multimodal features. The following subsections provide a detailed explanation of each module within the BDGF framework.

\begin{figure*}[thpb]
	\centering
	\includegraphics[width=\linewidth]{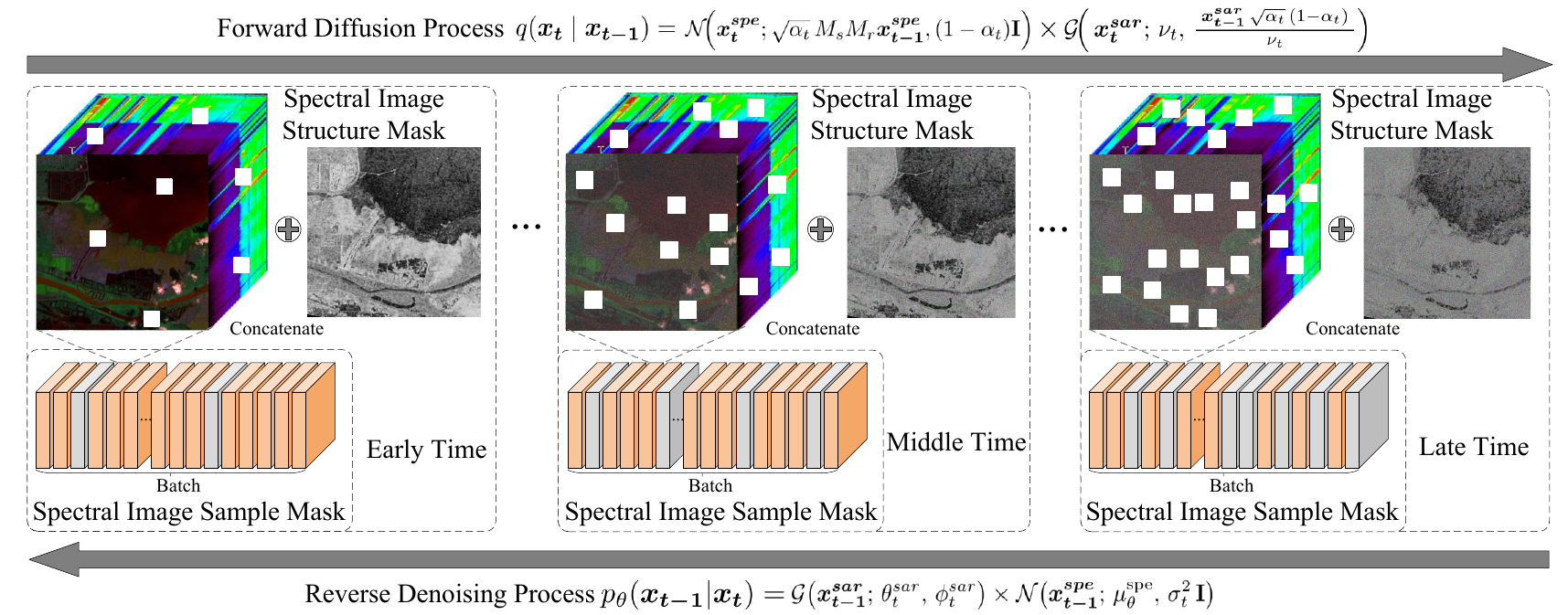}
	\caption
	{Structure of the adaptive modality masking strategy. In the forward diffusion process, the strategy consists of adding an iteration-varying structure mask and sample mask to the spectral image, while adding noise to the multimodal data.}
	\label{diff}
\end{figure*}

\subsection{Adaptive Modality Masking-Based DDPMs}
DDPMs excel at extracting noise-reduced and robust representations that capture complex data distributions, which are beneficial for feature extraction and classification. To fully exploit the advantages of DDPMs, we improve the model with respect to remote sensing image noise, multimodal fusion, and modality imbalance. The overall structure is shown in Fig. \ref{diff}.

Let us focus the attention on the most typical information source in remote sensing, i.e. SAR and spectral images. SAR images are severely disturbed by speckle noise \cite{lee1981speckle}, which makes it difficult to extract discriminative features. For an $L$-look SAR image, speckle noise is typically modeled as multiplicative noise $n$. With mean $I$ and variance $1/L$, the gamma-distributed probability density function is defined as:
\begin{align}
	p_{sar}(n) = {(\frac{L}{I})}^L \frac{ n^{L-1}}{\Gamma(L)} \exp(-\frac{Ln}{I}),
	\label{3-1-1}
\end{align}
where $\Gamma(\cdot)$ denotes the gamma function. In contrast, noise in spectral images is typically modeled as an additive process following a Gaussian distribution. In this context, the additive noise $n$ is assumed to have zero mean and variance $\sigma^2$, with the probability density function given by:
\begin{align}  
	p_{spe}(n) = \frac{1}{\sqrt{2\pi\sigma^2}} \exp\left(-\frac{n^2}{2\sigma^2}\right).
	\label{3-1-2}
\end{align} 

An advantage of merging SAR and spectral images is the reduction in computational complexity during pre-training \cite{10314566}. However, spectral images, which contain more discriminative information, tend to dominate the optimization process, causing DDPMs to gradually overlook the complementary information provided by SAR images.

Inspired by \cite{7169562,10694738}, we aim to prevent DDPMs from over-focusing on spectral images. Unlike reconstruction tasks that add large blocks of masks from a spatial or spectral perspective \cite{10216780,liu2024hybrid}, our strategy dynamically employs a sample mask $m_{s}$ and a structure mask $m_{r}$ for spectral images $\boldsymbol{x}^{spe}_m$. The sample mask $m_{s}$ reduces the proportion of the dominant modality in the batch, while $m_{r}$ randomly masks a certain proportion of the image using a minimal $1\times1\times1$ block. To dynamically suppress the dominant modality, the mask ratio is continuously increased during the iterative process. Assuming that $epoch \in [0, 1]$ represents the training progress, the mask generation can be expressed as:
\begin{equation}
	m \sim Bernoulli (\frac{\exp({epoch}) - \exp({-epoch})}{{\exp({epoch}) + \exp({-epoch})}}),
	\label{3-1-3}
\end{equation}
where $Bernoulli$ is Bernoulli distribution and $\exp$ is an exponential operation. This soft distribution allows unbiased and element-wise random masking based on the iteration progress. Based on (\ref{3-1-1})-(\ref{3-1-3}), we add multiplicative speckle noise to the SAR image $\boldsymbol{x}_0^{sar}$ and Gaussian noise to the spectral image $\boldsymbol{x}_0^{spe}$. Then, we merge them along the channel dimension. (\ref{2-1-1}) can be transformed into:
\begin{equation}
	\begin{aligned}
		\begin{cases}
			\boldsymbol{x}_t
			= \sqrt{\bar\alpha_t}
			\begin{bmatrix}
				\prod_{i=1}^{t}n_i^s \odot \boldsymbol{x}_0^{sar} \\
				M_{s}M_{r} \boldsymbol{x}_0^{spe} 
			\end{bmatrix}
			\;+\;
			\sqrt{1-\bar\alpha_t}
			\begin{bmatrix}
				0 \\
				\epsilon_p
			\end{bmatrix},\\
			\bar{\alpha}_t =   \prod_{i=1}^{t}n_i\alpha_i, M = \mathrm{diag}(m),
			n_i^s \,\sim\, p_{\rm sar},  \epsilon_p\,\sim\, p_{\rm spe}.
		\end{cases}
	\end{aligned}
	\label{7}
\end{equation}

Given $v_t$ and $\mathcal{G}$ as SAR Gamma shape and distribution respectively, the transfer probability of forward diffusion process (\ref{2-1-2}) can be expressed as:
\begin{equation}
	\begin{aligned}
		\begin{cases}
			q(\boldsymbol {x}_t \mid \boldsymbol {x}_{t-1})
			=
			q^{sar}\bigl(\boldsymbol{x}_t^{sar}\mid \boldsymbol{x}_{t-1}^{sar}\bigr)
			\times
			q^{spe}\bigl(\boldsymbol{x}_t^{spe}\mid \boldsymbol{x}_{t-1}^{spe}\bigr),
			\\[5pt]
			q^{spe}\bigl(\boldsymbol{x}_t^{spe}| \boldsymbol{x}_{t-1}^{spe}\bigr)
			=
			\mathcal N\!\Bigl(
			\boldsymbol{x}_t^{spe};
			\sqrt{\alpha_t}\,M_sM_r\boldsymbol{x}_{t-1}^{spe},
			(1-\alpha_t)\mathbf{I}
			\Bigr),\\[5pt]
			q^{sar}\bigl(\boldsymbol{x}_t^{sar}| \boldsymbol{x}_{t-1}^{sar}\bigr)
			=
			\mathcal{G}\!\Bigl(\,
			\boldsymbol{x}_t^{sar};\,
			\nu_t,\,
			\frac{\;\boldsymbol{x}_{t-1}^{sar}\,\sqrt{\alpha_t}\,(1-\alpha_t)\;}{\nu_t}
			\Bigr).	
		\end{cases}
	\end{aligned}
	\label{3-1-4}
\end{equation}

Correspondingly, the transfer probability of reverse diffusion process (\ref{2-1-3}) can be expressed as:
\begin{equation}
	\label{eq:reverse-single-net}
	\begin{aligned}
		p_\theta(\boldsymbol{{x}}_{t-1}|\boldsymbol{x}_t)
		=
		\mathcal G\bigl(
		\boldsymbol{x}_{t-1}^{sar};\,
		\theta_t^{sar},\,\phi_t^{sar}
		\bigr)
		\times
		\mathcal N\bigl(\boldsymbol{x}_{t-1}^{spe};\,
		\mu_\theta^{\rm spe},\,\sigma_t^2\,\mathbf I
		\bigr),
		\\
	\end{aligned}
\end{equation}
where $\theta_t^{\rm sar}$ and $\phi_t^{sar}$ are the Gamma shape and scale, respectively. $\mu_\theta^{\rm spe}$ and $\sigma_t^2$ are mean and variance similar to (\ref{2-1-3}). The denoising network updates its parameters via gradient descent. The pseudo-code is shown in Algorithm \ref{alg}.

%The noise predicted by the model is $\epsilon_\theta\bigl([\;\boldsymbol{x}_t^{ sar},\,\boldsymbol{x}_t^{spe}],\,t\bigr)$.
\begin{algorithm}
	\caption{Adaptive Modality Masking Strategy}
	\begin{algorithmic}[1]  % 显示行号（可选）
		\Require Original data $\boldsymbol{x}_0^{\rm sar}, \boldsymbol{x}_0^{\rm spe}$; current training progress $epoch \in [0, 1]$; diffusion schedule $\bar{\alpha}_t$.
		
		\State \textbf{Dynamic Mask Generation:}
		\State Compute mask probability: $p \leftarrow \frac{\exp(epoch) - \exp(-epoch)}{\exp(epoch) + \exp(-epoch)}$
		\State Compute sample and structure masks: $m_s, m_r$
		\State Construct diagonal mask matrices: $M_s = \mathrm{diag}(m_s)$, $M_r = \mathrm{diag}(m_r)$
		
		\State \textbf{Noisy Multimodal Data Generation:}
		\State Sample time step $t$
		\State Sample multiplicative speckle noise: $n^s \sim p_{sar}$
		\State Sample Gaussian noise: $\epsilon_p \sim \mathcal{N}(0, \mathbf{I})$
		
		\State Construct multimodal noisy state $\boldsymbol{x}_t$:
		\State $\boldsymbol{x}_t \leftarrow
		\begin{bmatrix}
			\sqrt{\bar\alpha_t} (\prod_{i=1}^{t}n_i^s \odot \boldsymbol{x}_0^{sar}) \\
			\sqrt{\bar\alpha_t} (M_{s}M_{r} \boldsymbol{x}_0^{spe}) + \sqrt{1-\bar\alpha_t} \epsilon_p
		\end{bmatrix}$
		
		\State \textbf{Diffusion Process and Noise Prediction:}
		\State Generate forward diffusion:
		\State $q(\boldsymbol {x}_t \mid \boldsymbol {x}_{t-1}) =
		q^{sar}(\boldsymbol{x}_t^{sar}\mid \boldsymbol{x}_{t-1}^{sar})
		\times
		q^{spe}(\boldsymbol{x}_t^{spe}\mid \boldsymbol{x}_{t-1}^{spe})$
		
		\State Generate reverse denoising:
		\State $p_\theta(\boldsymbol{{x}}_{t-1}|\boldsymbol{x}_t) =
		\mathcal G(
		\boldsymbol{x}_{t-1}^{sar},
		\theta_t^{sar},\phi_t^{sar})
		\times
		\mathcal N(\boldsymbol{x}_{t-1}^{spe},
		\mu_\theta^{\rm spe},\sigma_t^2\mathbf I)$
		
		\State Predict noise and update parameters.
	\end{algorithmic}
	\label{alg}
\end{algorithm}

\begin{figure}[ht]
	\centering
	\includegraphics[width=\linewidth]{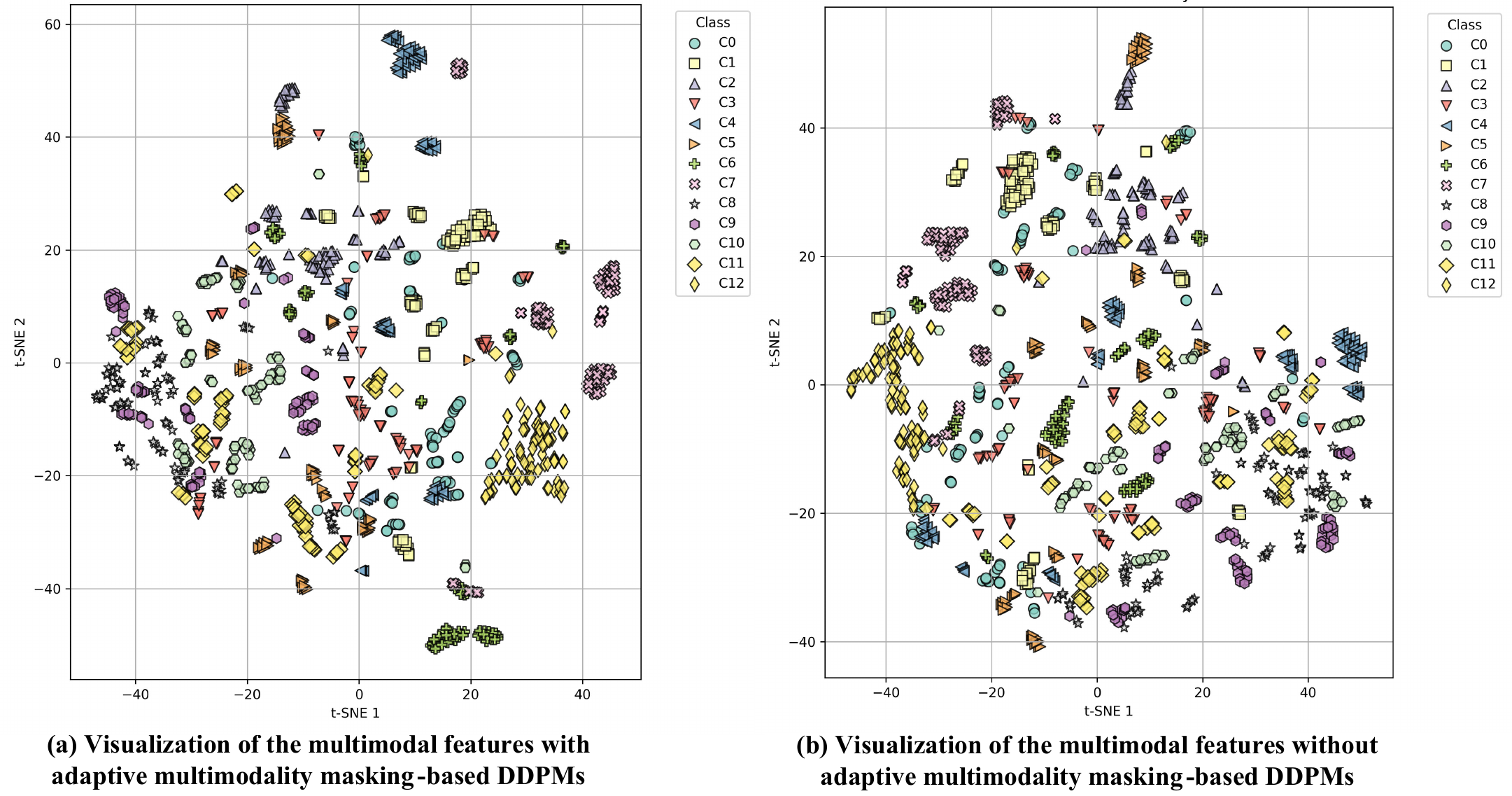}
	\caption
	{2D t-SNE embeddings of diffusion feature distribution on the LCZ HK dataset.}
	\label{dif_vis}
\end{figure}

Fig. {\ref{dif_vis}} visualizes the distribution effect of the features obtained by the proposed method on the LCZ HK dataset. The t-SNE execution details are the same as in Section \ref{visualization}. Across categories the adaptive multimodality masking (a) produces more uniform, consolidated class clusters and reduces dominance and fragmentation seen in the original model.For example, Class 6 forms a tight cluster in (a) but is scattered and intermixed with other classes in (b).

\subsection{Diffusion Features Guidance}

To leverage the multimodal data distribution extracted through the diffusion process, the diffusion features guide the extraction of diverse features in accordance with the characteristics of each branch. CNN-based architectures capture local features using small convolution kernels, whereas transformers and Mamba models extract global features via attention mechanisms and state space models. Mamba is particularly well-suited for tasks involving long sequence features \cite{yu2024mambaout}. Section \ref{visualization} presents a detailed visualization analysis of the complementarity of different features.

%\begin{figure}[ht]
%	\centering
%	\includegraphics[width=0.5\linewidth]{fig4.pdf}
%	\caption
%	{Flowchart of diffusion features guide CNN-based network.}
%	\label{cnn}
%\end{figure}
\subsubsection{Local Feature Guidance} 
CNN networks extract local information that differs significantly from diffusion features. Therefore, the diffusion data distribution is deeply integrated into feature generation via a feature fusion approach. Given the spectral image $\boldsymbol{x}^{spec}$, SAR image $\boldsymbol{x}^{sar}$, and diffusion features $\boldsymbol{f}^{dif}$ as inputs, the process is expressed as follows:
\begin{equation}
	\begin{aligned}  
		\boldsymbol{f}^1_{cnn} &=  \beta \cdot w_1^{3d} \boldsymbol{x}^{spec} + (1-\beta) \cdot w_1^{2d} \boldsymbol{f}^{dif},\\
		\boldsymbol{f}^2_{cnn} &=  \gamma \cdot w_2^{3d} \boldsymbol{f}^1_{cnn}  + (1-\gamma) \cdot w_2^{2d} \boldsymbol{f}^{dif},\\
		\boldsymbol{f}^3_{cnn} &=  \alpha \cdot w_4^{2d} \boldsymbol{x}^{sar} + (1-\alpha) \cdot w_3^{2d} \boldsymbol{f}^{dif},
		\label{3-2-1}
	\end{aligned}
\end{equation}
where $w^{3d}$ and $w^{2d}$ denote three-dimensional and two-dimensional convolution operations, respectively. The intermediate feature vectors $\boldsymbol{f}^1_{cnn}$, $\boldsymbol{f}^2_{cnn}$, and $\boldsymbol{f}^3_{cnn}$ are produced during the network's processing, and $\alpha$, $\beta$, and $\gamma$ are trainable scalar parameters. The output of the feature fusion-based CNN module is given by:
\begin{align}  
	\boldsymbol{f}_{cnn} = w_6^{2d} (w_3^{3d} \boldsymbol{f}_{cnn}^2 + w_5^{2d} \boldsymbol{f}^3_{cnn}).	
	\label{3-2-2}
\end{align}

\subsubsection{Global Feature Guidance} 
\begin{figure}[ht]
	\centering
	\includegraphics[width=0.65\linewidth]{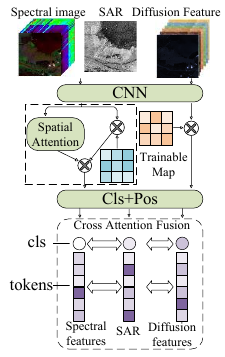}
	\caption
	{Illustration of global feature guidance structure.}
	\label{trans}
\end{figure}

The transformer exploits attention mechanisms to extract global information, a key aspect of feature diversity. As shown in Fig. \ref{trans}, the transformer-based network comprises trainable mapping tensors for preliminary processing of multimodal data and a cross-attention fusion module to facilitate global feature interaction. Given the spectral image $\boldsymbol{x}^{spec}$, the output $\boldsymbol{f}^{spec}_{trans}$ of spatial attention and trainable mapping is computed as:
\begin{equation}
	\begin{aligned}  
		\boldsymbol{f}^{spec}_{trans} =  Soft \cdot (\boldsymbol{W}_1^{att} \cdot w_7^{2d} \boldsymbol{x}^{spec}) \cdot \boldsymbol{W}_1^{map} \cdot \boldsymbol{x}^{spec},
		\label{3-2-3}
	\end{aligned}
\end{equation}
where $\boldsymbol{W}_1^{att}$ and $Soft$ denote a trainable tensor and an activation function, respectively, to obtain the spatial importance. The tensor $\boldsymbol{W}_1^{map}$ maps features to a common dimension, and $w_7^{2d}$ represents a two-dimensional convolution operation. Similarly, the intermediate feature $\boldsymbol{f}^{sar}_{trans}$ is obtained from the SAR image $\boldsymbol{x}^{sar}$. The diffusion features, after tensor mapping, are given by:
\begin{equation}
	\begin{aligned}  
		\boldsymbol{f}^{dif}_{trans} =  \boldsymbol{W}_2^{map} \cdot \boldsymbol{f}_{dif},
		\label{3-2-4}
	\end{aligned}
\end{equation}
where $\boldsymbol{W}_2^{map}$ is a mapping tensor for processing diffusion features.

Furthermore, inspired by \cite{9772757,9999457}, cross-attention fusion is employed to share abstract classification information among spectral, SAR, and diffusion features. Taking spectral features as an example, the $\boldsymbol{Q}$, $\boldsymbol{K}$, and $\boldsymbol{V}$ in the self-attention mechanism are computed as:
\begin{equation}
	\begin{aligned}  
		\boldsymbol Q, \boldsymbol K, \boldsymbol V = FC \cdot \boldsymbol{f}^{spec}_{trans},
		\label{3-2-5}
	\end{aligned}
\end{equation}
where $FC$ represents linear layers that generate attention vectors with the same dimensionality as the input. The self-attention mechanism then produces a new feature vector:
\begin{equation}
	\begin{aligned}  
		\boldsymbol{F}^{spec}= \boldsymbol V \cdot Soft \left(\frac{\boldsymbol Q\cdot \boldsymbol K^T}{\sqrt{d_k}}\right) +\boldsymbol {f^{spec}_{trans}},
		\label{3-2-6}
	\end{aligned}
\end{equation}
where $d_k$ denotes the dimension of $\boldsymbol{K}$. Similarly, self-attention is applied to obtain $\boldsymbol{F}^{sar}$ and $\boldsymbol{F}^{dif}$ for the SAR and diffusion features, respectively. These vectors comprise both class tokens and patch tokens. Spectral and SAR images can be expressed as $\boldsymbol{F}^{spec} = \boldsymbol{F}^{spec}_{cls} \cup \boldsymbol{F}^{spec}_{tok}$ and $\boldsymbol{F}^{sar} = \boldsymbol{F}^{sar}_{cls} \cup \boldsymbol{F}^{sar}_{tok}$. 

Subsequently, the cross attention vectors are computed as:
\begin{equation}
	\begin{aligned}  
		\boldsymbol {Q}^{spec} = \boldsymbol{W}_Q\boldsymbol{F}^{spec}_{cls}, \;
		\boldsymbol {K}^{sar} =\boldsymbol{W}_K\boldsymbol{F}^{sar}_{tok}, \;
		\boldsymbol {V}^{sar} =\boldsymbol{W}_V\boldsymbol{F}^{sar}_{tok},
		\label{3-2-7}
	\end{aligned}
\end{equation}
where $\boldsymbol{W}_Q$, $\boldsymbol{W}_K$, and $\boldsymbol{W}_V$ are trainable weight tensors. The new spectral feature is then computed as:
\begin{equation}
	\begin{aligned}  
		\boldsymbol{F}^{spec}_{trans} = 
		(\boldsymbol {V}^{sar} \cdot Soft \frac{\boldsymbol {Q}^{spec}\cdot(\boldsymbol{K}^{sar})^{T} }{\sqrt{d_k^{sar}}}  
		+ \boldsymbol {F}^{spec}_{trans} ) \cup \boldsymbol{F}^{spec}_{tok},
		\label{3-2-8}
	\end{aligned}
\end{equation}
where $d_k^{sar}$ represents the dimension of $\boldsymbol{K}^{sar}$. Similarly, $\boldsymbol{F}^{sar}_{trans}$ is computed. Finally, after combining $\boldsymbol{F}^{spec}_{trans}$ and $\boldsymbol{F}^{sar}_{trans}$, the merged feature is processed with $\boldsymbol{F}^{dif}$ in a manner analogous to \eqref{3-2-7} and \eqref{3-2-8} to obtain $\boldsymbol{F}^{diff}_{trans}$. The final output of the transformer-based network is:
\begin{equation}
	\begin{aligned}  
		\boldsymbol{f}_{trans} = \boldsymbol{F}^{spectral}_{trans}+\boldsymbol{F}^{sar}_{trans}+\boldsymbol{F}^{diff}_{trans}.
		\label{3-2-9}
	\end{aligned}
\end{equation}

\subsubsection{Sequential Feature Guidance}
\begin{figure}[th]
	\centering
	\includegraphics[width=\linewidth]{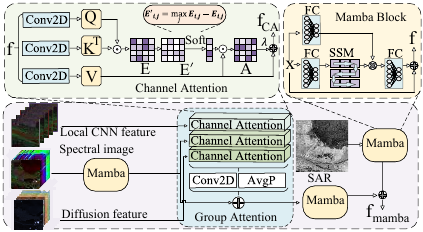}
	\caption
	{Flowchart of the proposed sequential feature guidance.}
	\label{mamb}
\end{figure}

Fig. \ref{mamb} illustrates the flowchart of the Mamba-based network, which incorporates Mamba blocks to extract long-sequence information and a group attention module to alleviate data imbalance.

Spectral images naturally yield longer sequences compared to SAR images. However, spectral redundancy can cause a modality imbalance that restricts the information extraction capability of Mamba. To address this limitation, we integrate local CNN features from \eqref{3-2-2} and global diffusion features, employing a balanced global-local group attention mechanism to handle redundant spectral dimensions.

Taking the spectral image $\boldsymbol{x^{spec}}$ as input, the operations in Mamba are defined as:
\begin{equation}
	\begin{aligned}
		\boldsymbol{F}^{spec}_{m}&=(SSM \cdot FC \cdot \boldsymbol{x}^{spec}) \otimes (FC \cdot \boldsymbol{x}^{spec}),\\
		\boldsymbol{f}^{spec}_{m}&=(FC \cdot \boldsymbol{F}^{spec}_{m}) \oplus \boldsymbol{x}^{spec},
		\label{3-2-13}
	\end{aligned}
\end{equation}
where $\boldsymbol{F}^{spec}_{m}$ is the intermediate feature produced by the Mamba blocks, $\boldsymbol{f}^{spec}_{m}$ is the final output, $SSM$ denotes the SSM module with the selective scan mechanism, and $\otimes$, $\oplus$ denote appropriate fusion operations. Similarly, the SAR image $\boldsymbol{x}^{sar}$ yields features $\boldsymbol{f}^{sar}_{m}$.

Subsequently, $\boldsymbol{f}^{spec}_{m}$, $\boldsymbol{f}^{dif}$, and $\boldsymbol{f}_{cnn}$ from \eqref{3-2-2} are used to construct group attention. Assuming the input feature is $\boldsymbol{f}$, the attention mechanism is expressed as:
\begin{equation}
	\begin{aligned}
		\boldsymbol {Q}_{m}, \boldsymbol {K}_{m}, \boldsymbol {V}_{m} &= w_8^{2d}(\boldsymbol{f}), w_9^{2d}(\boldsymbol{f}), w_{10}^{2d}(\boldsymbol{f}),\\
		\boldsymbol{E}_{m} &= \boldsymbol {Q}_{m} \cdot {\boldsymbol {K}_{m}}^{T},
		\label{3-2-14}
	\end{aligned}
\end{equation}
where $\boldsymbol {Q}_{m}$, $\boldsymbol {K}_{m}$, $\boldsymbol {V}_{m}$ and $\boldsymbol{E}_{m}$ are the vectors in the attention mechanism. $w_8^{2d}$, $w_9^{2d}$, and $w_{10}^{2d}$ denote two-dimensional convolution operations. For each $i$ and $j$ dimension of tensor $\boldsymbol{E}_{m}$, we update it to obtain the attention score $\boldsymbol{A}_m$ by taking the maximum along the $j$ dimension and expanding its shape:
\begin{equation}
	\begin{aligned}
		\boldsymbol{E'}_{i,j} &= \max_{j}\boldsymbol{E}_{i,j} - \boldsymbol{E}_{i,j},\\
		\boldsymbol{A}_{m} &= \boldsymbol {V}_{m} \cdot Soft (\boldsymbol{E'}_{m}),
		\label{3-2-15}
	\end{aligned}
\end{equation}
where $\boldsymbol{E'}_{m}$ denotes the updated tensor. The output of the single-channel attention is:
\begin{equation}
	\begin{aligned}
		\boldsymbol{f}_{CA} = \lambda \boldsymbol{A}_{m} + \boldsymbol {V}_{m},
		\label{3-2-16}
	\end{aligned}
\end{equation}
where $\lambda$ as a trainable scaling parameter. Similarly, we obtain outputs $\boldsymbol{f}^{spec}_{CA}$, $\boldsymbol{f}^{dif}_{CA}$, and $\boldsymbol{f}_{CA}^{cnn}$. The output of the group attention module is given by:
\begin{equation}
	\begin{aligned}
		\label{3-2-17}
		\boldsymbol{F}^{spec}_{GA} = w_{11}^{2d}\left(\boldsymbol{f}^{spec}_{CA}, \boldsymbol{f}^{dif}_{CA}, \boldsymbol{f}_{CA}^{cnn}\right)\cdot \boldsymbol{f}^{spec}_{m}+\boldsymbol{f}^{spec}_{m}.
	\end{aligned}
\end{equation}
%where $\boldsymbol{f^{spec}_{m}}$ represents the output from the Mamba blocks. 

Finally, based on $\boldsymbol{f}^{sar}_{m}$ in \eqref{3-2-13} and $\boldsymbol{F}^{spec}_{GA}$ in \eqref{3-2-17}, the output of the Mamba-based network $\boldsymbol{f}_{mamba}$ is obtained.

\subsection{Mutual Learning Module}

The mutual learning module promotes collaboration among sub-networks and enhances the fusion of diverse features \cite{10122197}. The experiments in Section \ref{visualization} visualize the effect of this strategy. This module uses KL divergence to align the entropy and feature similarity of paired networks.

Taking features $\boldsymbol{f}_{cnn}$ and $\boldsymbol{f}_{trans}$ as an example, their feature similarity is defined by cosine similarity:
	\begin{equation}
		\begin{aligned}
			\label{3-3-1}
			{Sim}(\boldsymbol {f}_{cnn}, \boldsymbol {f}_{trans}) = \frac{\boldsymbol {f}_{cnn} \cdot \boldsymbol {f}_{trans}}{\|\boldsymbol {f}_{cnn}\|\, \|\boldsymbol {f}_{trans}\|}.
		\end{aligned}
	\end{equation}
	
Each sub-network produces a classification probability $\boldsymbol{z}$, from which the categorical probability distribution is derived as:
	\begin{equation}
		\begin{aligned}
			\label{3-3-2}
			\boldsymbol{p}^{(i)}(j) = \frac{\exp(\boldsymbol{z}_j^{(i)})}{\sum_{k=1}^C \exp(\boldsymbol{z}_k^{(i)})},\\
			H^{(i)} = -\sum_{j=1}^C \boldsymbol{p}^{(i)}(j) \log \boldsymbol{p}^{(i)}(j),
		\end{aligned}
	\end{equation}
where $i \in B$ and $j \in C$ denote the sample and category indices, respectively, with $B$ samples for a batch and $C$ classes. $H^{(i)}$ represents the entropy. Thus, the classification entropies are denoted as $H_{cnn}$, $H_{trans}$, and $H_{mamba}$.
	
The similarity and entropy from \eqref{3-3-1} and \eqref{3-3-2} jointly determine the temperature in KL divergence:
	\begin{equation}
		\begin{aligned}
			\begin{cases}
				temp =  \left({Sim}(\boldsymbol {f}_{cnn}, \boldsymbol {f}_{trans}), H_{cnn}+H_{trans}\right),\\
				T = \ln\left(1 + \exp(FC\cdot {temp})\right) + 10^{-6}.
			\end{cases}
		\end{aligned}
		\label{3-3-3}
	\end{equation}
	
Integrating this adaptive temperature into the KL divergence, the mutual learning loss between the CNN and transformer networks is defined as:
	\begin{equation}
		\begin{aligned}
			\label{3-3-4}
			L_{\text{kl}}^{ct} =
			T^2 \cdot \frac{1}{B} \sum_{i=1}^{B} \sum_{j} \boldsymbol{p}_{trans}^{(i)}(j) \Bigl[\log \boldsymbol{p}_{trans}^{(i)}(j) - \log \boldsymbol{p}_{cnn}^{(i)}(j)\Bigr].
		\end{aligned}
	\end{equation}
	
Similarly, the mutual learning losses $L_{\text{kl}}^{cm}$ and $L_{\text{kl}}^{tm}$ are computed for the other paired sub-networks. Given cross-entropy loss function $L_{\text{ce}}$, the final loss is expressed as:
	\begin{equation}
		\begin{aligned}
			\label{3-3-5}
			\begin{cases}
				\boldsymbol {z}_{\text{total}} = FC \cdot (\boldsymbol{f}_{trans}, \boldsymbol{f}_{cnn},\boldsymbol{f}_{mamba}),\\
				L_{\text{total}} = L_{\text{ce}} (\boldsymbol {z}_{\text{total}}) + L_{\text{kl}}^{ct}+L_{\text{kl}}^{cm}+L_{\text{kl}}^{tm}.
			\end{cases}
		\end{aligned}
	\end{equation}

\section{Experimental Results and Discussion}
\definecolor{m1}{HTML}{FF0000}
\definecolor{m2}{HTML}{00FF00}
\definecolor{m3}{HTML}{0000FF}
\definecolor{m4}{HTML}{FFFF00}
\definecolor{m5}{HTML}{00FFFF}
\definecolor{m6}{HTML}{FF00FF}
\definecolor{m7}{HTML}{C0C0C0}
\definecolor{m8}{HTML}{808080}
\definecolor{m9}{HTML}{800000}
\definecolor{m10}{HTML}{808000}
\definecolor{m11}{HTML}{008000}
\definecolor{m12}{HTML}{800080}
\definecolor{m13}{HTML}{008080}
\definecolor{m14}{HTML}{0000FF}
\definecolor{m15}{HTML}{FFA500}

Four multimodal remote sensing datasets are used to evaluate the classification performance of the proposed BDGF. In this section, we first introduce the datasets and evaluation criteria, followed by detailed ablation experiments. Next, we visualize feature complementarity of the two modules on the LCZ HK dataset. Finally, we compare BDGF with state-of-the-art methods and discuss its transferability and computational complexity.

We compare BDGF with several state-of-the-art multimodal remote sensing classification methods, including CNN-based networks AsyFFNet \cite{9716784} and CALC \cite{lu2023coupled}, and the pre-training method SS-MAE \cite{10314566}. In addition, we include four classification methods that focus on multi-branch and multi-scale networks: Fusion-HCT \cite{9999457}, MACN \cite{li2023mixing}, NCGLF \cite{tu2024ncglf2}, and UACL \cite{10540387}. Two advanced multi-scale Mamba-based methods, HLMamba \cite{10679212} and MSFMamba \cite{10856240}, are also evaluated.

\subsection{Description of Datasets}
\begin{figure*}[htbp]
	\includegraphics[width=0.8\linewidth]{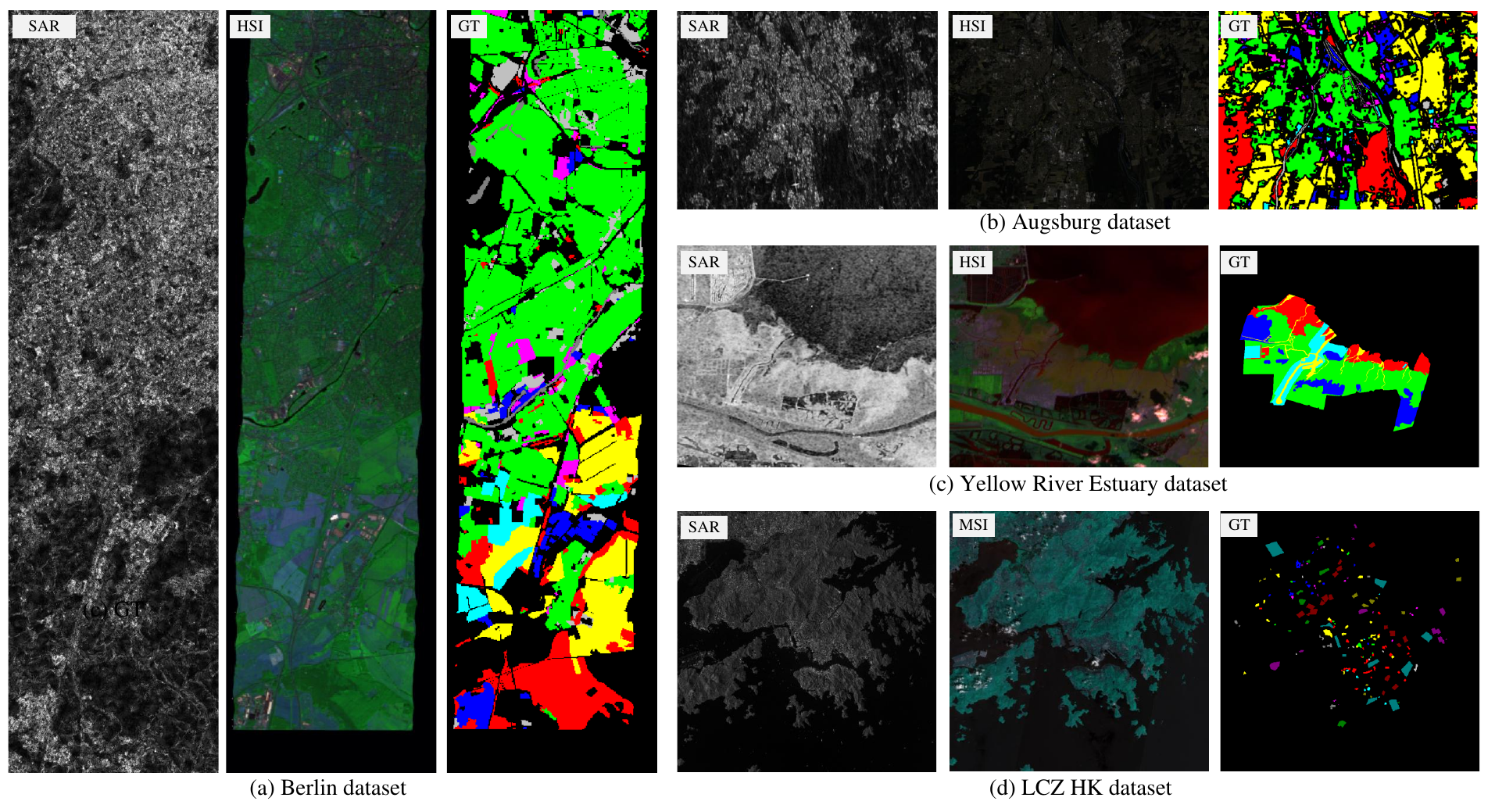}
	\centering
	\caption{Multimodal remote sensing datasets. (a) Berlin dataset. (b) Augsburg dataset. (c) Yellow River Estuary dataset. (d) LCZ HK dataset} 
	\label{dataset}
\end{figure*}
\begin{table*}[ht]
	\centering
	\caption{Land-cover classes and related numbers of samples in the four considered datasets}
	\label{table:samples}
	\renewcommand{\arraystretch}{1.2}
	\setlength{\tabcolsep}{4pt}
	\resizebox{0.8\textwidth}{!}{
		\begin{tabular}{|c|c|c|}
			\hline
			\begin{tabular}[t]{@{}cllc@{}}
				\multicolumn{4}{c}{\textbf{Augsburg (HSI+SAR)}} \\
				\hline
				\textbf{No.} & \textbf{Color} & \textbf{Name} & \textbf{Numbers} \\
				\hline
				
				1 & \cellcolor{m1}  & Forest              & 13507 \\
				2 & \cellcolor{m2} & Residential Area    & 30329 \\
				3 & \cellcolor{m3} & Industrial Area     & 3851  \\
				4 &\cellcolor{m4}  & Low Plants          & 26857 \\
				5 & \cellcolor{m5}  & Allotment           & 575  \\
				6 & \cellcolor{m6}  & Commercial Area     & 1645  \\
				7 & \cellcolor{m7}  & Water               & 1530  \\
				
				\textbf{Total} & & & 78294 \\
				\hline
				\multicolumn{4}{c}{\textbf{Yellow River Estuary (HSI+SAR)}} \\
				\hline
				\textbf{No.} & \textbf{Color} & \textbf{Name} & \textbf{Numbers} \\
				\hline
				1 & \cellcolor{m1}  & Spartina Alterniflora           & 39784 \\
				2 & \cellcolor{m2}  & Suaeda Salsa    & 118213 \\
				3 & \cellcolor{m3}  & Tamarix Forest     & 35216 \\
				4 & \cellcolor{m4}  & Tidal Creek          & 15673 \\
				5 & \cellcolor{m5}  & Mudflat                & 24592 \\
				
				\textbf{Total} & & & 233478 \\
				\hline
			\end{tabular}
			&
			\begin{tabular}[t]{@{}cllc@{}}
				\multicolumn{4}{c}{\textbf{Berlin (HSI+SAR)}} \\
				\hline
				\textbf{No.} & \textbf{Color} & \textbf{Name} & \textbf{Numbers} \\
				\hline
				1 & \cellcolor{m1}  & Forest              & 54954 \\
				2 & \cellcolor{m2} & Residential Area    & 268642 \\
				3 & \cellcolor{m3} & Industrial Area     & 19566  \\
				4 &\cellcolor{m4}  & Low Plants          & 59282 \\
				5 & \cellcolor{m5}  &Soil 					& 17426\\
				6 & \cellcolor{m6}  & Allotment           & 13305 \\
				7 & \cellcolor{m7}  & Commercial Area     & 24824  \\
				8 & \cellcolor{m8}  & Water               & 6672  \\
				&&&\\
				&&&\\
				&&&\\
				&&&\\
				&&&\\
				&&&\\
				&&&\\
				
				\textbf{Total} & & & 464671\\
				\hline
			\end{tabular}
			&
			% Table 2: Yellow Dataset
			\begin{tabular}[t]{@{}cllc@{}}
				\multicolumn{4}{c}{\textbf{LCZ HK (MSI+SAR)}} \\
				\hline
				\textbf{No.} & \textbf{Color} & \textbf{Name} & \textbf{Numbers} \\
				\hline
				1  & \cellcolor{m1} & Compact High-rise & 631 \\
				2  & \cellcolor{m2} &  Compact Mid-rise              & 179  \\
				3  & \cellcolor{m3}  &Compact Low-rise & 326  \\
				4  & \cellcolor{m4}  & Open High-rise           & 673  \\
				5  & \cellcolor{m5}  & Open Mid-rise & 126    \\
				6  & \cellcolor{m6}  & Open Low-rise & 120  \\
				7  &\cellcolor{m7}  &Large Low-rise  & 137  \\
				8  & \cellcolor{m8} &  Heavy Industry       & 219  \\
				9  & \cellcolor{m9} &  Dense Trees & 1616  \\
				10 & \cellcolor{m10}  &   Scattered Trees          & 540   \\
				11 &\cellcolor{m11}  &   Bush and Scrub         & 691  \\
				12 & \cellcolor{m12}  & Low Plants          & 985  \\
				13 &\cellcolor{m13} &  Water        & 2603  \\
				&&&\\
				&&&\\
				
				\textbf{Total} & & & 8846\\
				\hline
			\end{tabular}
			
		\end{tabular}
	}
\end{table*}

%\subsubsection{Houston 2013 dataset (HSI+LiDAR)}The Houston2013 dataset, captured by the compact airborne spectrographic imager sensor in 2012 over the University of Houston, provides a detailed view of the urban campus and its surroundings. This dataset consists of HSI and LiDAR data, with each information source having a spatial resolution of 2.5 meters and dimensions of 349$\times$1905 pixels. The HSI comprises 144 bands spanning the wavelength range from 0.38 to 1.05$\mu$m, while the LiDAR data offer elevation measurements. The dataset is divided into the 15 distinct land-cover classes presented in Table \ref{table:samples} (a) together with the number of labeled samples. A visual representation of the dataset, including a pseudo-color composite of the HSI and a grayscale image of the LiDAR data, is displayed in Fig. \ref{dataset}, alongside the ground-truth map.

\subsubsection{Berlin dataset (HSI+SAR)}
The Berlin dataset provides a comprehensive view of urban and rural regions in Berlin, Germany. It comprises HSI and SAR data, each with a spatial resolution of 30 meters and dimensions of 797$\times$220 pixels. The HSI data, collected by the HyMap sensor (simulated for the EnMAP satellite), consist of 244 spectral bands covering 400--2500 nm. The SAR data, captured by Sentinel-1, have been processed with SNAP for orbit correction, radiometric calibration, and speckle reduction. The dataset is divided into eight distinct land-cover classes, as detailed in Table \ref{table:samples}. A pseudo-color composite of the HSI, a grayscale SAR image, and the ground-truth map are presented in Fig. \ref{dataset} (a).

\subsubsection{Augsburg dataset (HSI+SAR)}
The Augsburg dataset captures a detailed rural landscape near Augsburg, Germany. It comprises a 332$\times$485 pixel HSI and a SAR image. The HSI, acquired by the HySpex sensor, covers 180 spectral bands from 400 to 2500 nm with a 30 m ground sampling distance. The SAR image, obtained by Sentinel-1 and preprocessed by the European Space Agency using the Sentinel Application Platform, is available in both dual-polarization (VV-VH) and single-look complex (SLC) formats. The dataset is categorized into seven land-cover classes at 30 m resolution. Fig. \ref{dataset} (b) visualizes the data through pseudo-color composites and a ground-truth map, while Table \ref{table:samples} summarizes the sample counts.

\subsubsection{Yellow River Estuary dataset (HSI+SAR)}
The Yellow River Estuary dataset \cite{gao2022fusion} provides a detailed perspective on wetland scenes in Shandong Province, China. The dataset, comprising 960$\times$1170 pixels with a spatial resolution of 30 meters, includes HSI and SAR data covering five land-cover classes. The HSI is acquired by the Advanced Hyperspectral Imager onboard the ZY1-02D satellite, covering 166 bands with spectral resolutions of 10 nm and 20 nm. Preprocessing of the HSI was performed with ENVI for radiometric and atmospheric correction. The SAR data were captured by Sentinel-1. Fig. \ref{dataset} (c) displays a pseudo-color composite of the HSI, a grayscale SAR image, and the ground-truth map, with sample details provided in Table \ref{table:samples}.

\begin{table*}[htbp]
	\centering
	\footnotesize
	\small
	\caption{OA (\%) , AA (\%) , and Kappa (\%)obtained in the ablation study on the four considered datasets (bold values are the best and underline values are the second)}
	\label{ablation}
	\renewcommand{\arraystretch}{1.5}
	\resizebox{\textwidth}{!}{
		\begin{tabular}{c|c|c|c|c|c|c|c|c|ccc|ccc|ccc|ccc}
			\hline
			\hline
			\multirow{2}{*}{Experiment Number} &\multirow{2}{*}{CNN} & \multirow{2}{*}{Trans} & \multirow{2}{*}{Mamba} & \multirow{2}{*}{Guide-CNN} & \multirow{2}{*}{Guide-Trans} & \multirow{2}{*}{Guide-Mamba} &\multirow{2}{*}{Mutual} &\multirow{2}{*}{Mask}&
			\multicolumn{3}{c|}{Augsburg dataset} & \multicolumn{3}{c|}{Berlin dataset} & \multicolumn{3}{c|}{Yellow River Estuary dataset} & \multicolumn{3}{c}{LCZ HK dataset} \\
			\cline{10-21}
			&& & &  &  &  &  & &OA & AA & Kappa & OA & AA & Kappa & OA & AA &Kappa & OA & AA & Kappa \\
			\hline
			1&\checkmark& & & \checkmark &  &  &  &\checkmark& \underline{93.23}&88.13&	\underline{90.41}	&	73.92&	76.53&	\underline{64.72}	&	74.48&	78.32&	66.09&		94.95&	95.26&	93.93\\
			2&&\checkmark & & &\checkmark  &  &  & \checkmark&92.02	&87.00&	88.79	&	69.60&	78.53&	57.50&		67.57&	67.35&	54.79	&	87.46&	88.81&	85.00 \\
			
			3&& &\checkmark & &    &\checkmark&  & \checkmark&91.02&	87.11&	87.42&		72.54&	78.17&	61.86&		74.42&	77.26&	64.58	&	92.52&	92.76&	91.03\\
			4&&\checkmark &\checkmark & &\checkmark  & \checkmark &\checkmark  & \checkmark&92.29&	88.06&	89.15&		68.80&	75.81&	57.50&	76.97	&78.03&	67.62	&	90.16&	90.98&	88.21\\
			5&\checkmark&&\checkmark & \checkmark & &\checkmark  &\checkmark & \checkmark&92.32&	89.36&	89.33	&	74.60&	78.96&	64.30&		78.98&	78.30&	70.15	&	95.12&	95.29&	94.14\\
			6&\checkmark&\checkmark & & \checkmark &\checkmark  &  &\checkmark & \checkmark&92.10&	89.76&	88.89	&	73.02&	78.52&	62.63	&	78.16&	77.84&	\underline{71.80}&		94.73&	95.02&	93.68\\
			7&\checkmark&\checkmark &\checkmark & &\checkmark  & \checkmark &\checkmark &\checkmark&92.93&	88.26&	90.18&		70.18&	76.57&	58.73&		77.73&	79.04&	67.33&94.40	&94.73&	93.29\\
			8&\checkmark&\checkmark &\checkmark & \checkmark & & \checkmark &\checkmark& \checkmark&92.50&	90.08&	89.48&		\underline{74.78}&	78.77&	64.49&		79.24&	79.66&	\pmb{72.06}	&	\underline{95.29}&	\textbf{95.80}&	\underline{94.35}
			\\
			9&\checkmark&\checkmark &\checkmark & \checkmark &\checkmark  & &\checkmark& \checkmark&92.78&	\pmb{90.46}&	89.84&		74.40&	78.72&	62.82	&	\underline{79.27}&	\underline{80.02}&	70.65&		94.74&	95.39&	93.69
			\\
			10&\checkmark&\checkmark &\checkmark & \checkmark &\checkmark  & \checkmark & & \checkmark&93.20&	89.93&	90.31	&	74.21&	\underline{78.98}&	64.20&	79.12&	79.35&	70.55&		94.62&	95.02&	93.55
			\\
			11&\checkmark&\checkmark &\checkmark & \checkmark &\checkmark  & \checkmark &\checkmark & &92.69&90.09	&89.70	&74.26	&76.70	&63.12	&78.03&80.35&69.37	&94.20		&95.38&	93.05
			\\
			12&\checkmark&\checkmark &\checkmark & \checkmark &\checkmark  & \checkmark &\checkmark & \checkmark&\pmb{93.57}&	\underline{90.12}&\pmb{90.93}	&	\pmb{75.11}&	\pmb{79.94}&	\pmb{64.73}	&	\pmb{79.55}&	\pmb{80.21}&	71.00&		\pmb{95.35}&	\underline{95.43}&	\pmb{94.42}
			\\
			\hline
			\hline
		\end{tabular}
	}
\end{table*}

\subsubsection{LCZ HK dataset (MSI+SAR)}
The Local Climate Zone Hong Kong (LCZ HK) dataset \cite{9174822} offers a comprehensive view of urban and rural areas in Hong Kong, China. It includes multispectral data collected by Sentinel-2 and SAR data from Sentinel-1. The MSI consists of ten spectral bands, resampled to a 100 m resolution, with a spatial size of 529$\times$528 pixels; the SAR image is downscaled to the same size. The dataset is divided into eight distinct land-cover classes as shown in Table \ref{table:samples}. Fig. \ref{dataset} (d) presents pseudo-color composites of the MSI and SAR data alongside the ground-truth map.

In our experiments, we employ four metrics to quantitatively assess classification performance: class-specific accuracy, overall accuracy (OA), average accuracy (AA), and the kappa coefficient (Kappa). The experiments are implemented using PyTorch and executed on an NVIDIA GeForce RTX 3090 (24 GB). For a fair comparison, following Spectraldiff \cite{10234379}, after pre-training, we select the single-step diffusion features after full down-sampling at time t=5 as input to the sub-network for classification. In addition, the denoising network also follows its U-Net structure but adds our masking strategy. Optimization is performed using the Adam algorithm with a learning rate of $4 \times 10^{-4}$, modulated by a MultiStepLR scheduler with a decay factor of 0.5. The patch size is 9 and the dimension of embedding features is 64. The implementation parameter details are in the publicly available code to ensure reproducibility \footnote{https://github.com/HaoLiu-XDU/BDGF}. All the comparison methods are executed under the same configurations. HSI and SAR datasets randomly use 100 labeled samples per class, while the MSI and SAR dataset randomly uses 50 labeled samples per class.

\subsection{Ablation Study}
To evaluate the effectiveness of the BDGF framework, we conduct a series of ablation experiments by selectively retaining key modules and sub-networks. Experiments 1–3 employ only a single network combined with diffusion features for feature extraction and classification to assess the contribution of each sub-network individually. In experiments 4–6, two networks are fused with diffusion features to evaluate their joint performance. To assess the impact of diffusion feature hierarchical guidance, experiments 7–9 are performed by removing the respective guidance branches. Finally, experiments 10 and 11 remove the adaptive modality mask strategy and the mutual learning module, respectively, to demonstrate their individual contributions. In Table \ref{ablation}, “CNN,” “Trans,” and “Mamba” denote networks based on CNN, transformer, and Mamba architectures, while “Guide-CNN,” “Guide-Trans,” and “Guide-Mamba” indicate the corresponding diffusion feature guidance modules. “Mutual” and “Mask” represent the mutual learning module and the adaptive modality mask strategy, respectively.

\begin{figure*}[htpb]
	\includegraphics[width=0.9\linewidth]{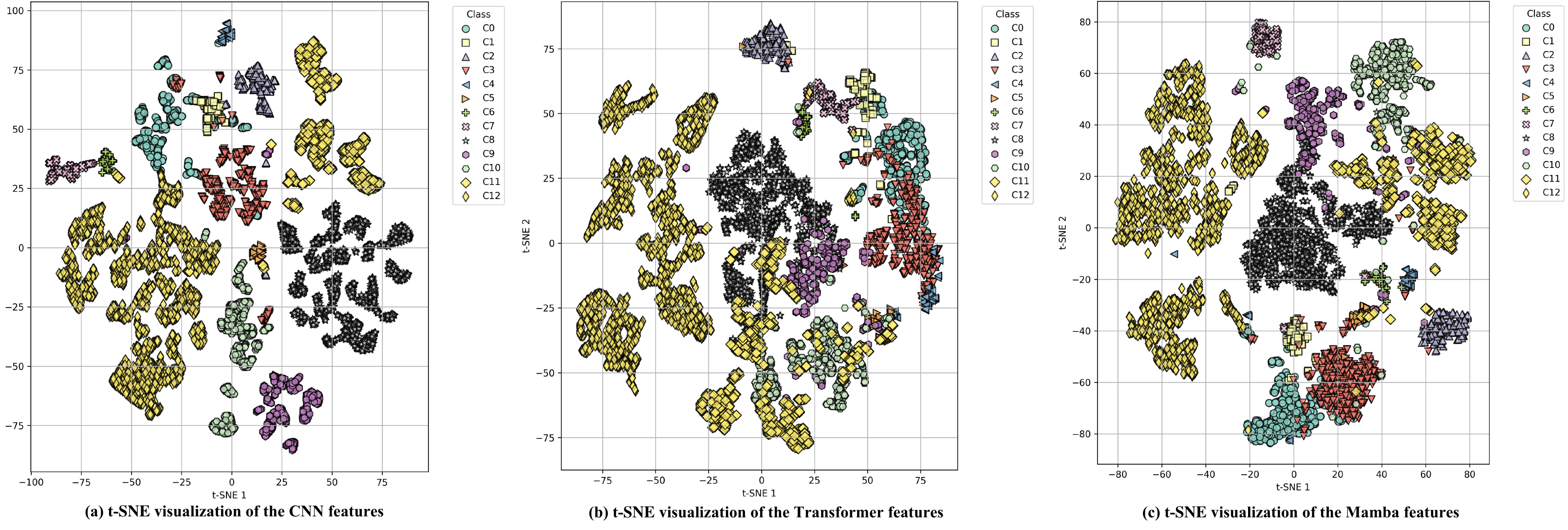}
	\centering
	\caption{2D t-SNE embeddings of per-branch features on the LCZ HK dataset. (a)–(c) represents features from the CNN, Transformer, and Mamba branches, respectively.}
	\label{9vis} 
\end{figure*}

%\begin{table}[hptb]\scriptsize
%	\renewcommand\arraystretch{2}
%	\centering
%	\caption{\\OA (\%) OBTAINED IN THE ABLATION STUDY ON THE FOUR CONSIDERED DATASETS}% 表格标题
%	\label{Table_ablation}
%	\setlength{\tabcolsep}{1.3mm}{
	%		\begin{tabular}{c|c|c|c|c|c} 
		%			\hline % 顶部线
		%			\hline
		%			Dataset&AFCM & HFGM &HFEST& SSAF& SSFin \\
		%			\hline
		%			Houston 2013  & 88.56 & 88.41   &   88.85&   89.02   &\pmb{89.19}     \\
		%			Yellow River Estuary  & 67.02&66.54&66.96&  67.00 &\pmb{67.56}  \\
		%			Augsburg  & 78.34&77.83&   79.88&  78.46 &\pmb{79.91}  \\
		%			LCZ HK  & 71.20 &   71.26 &71.60&   72.12 & \pmb{72.26}  \\
		%			
		%			\hline % 底部线
		%			\hline
		%			
		%			
		%		\end{tabular}
	%	}
%\end{table}

The experimental results are presented in Table \ref{ablation}. In general, the following conclusions can be drawn:
\begin{enumerate}
	\item Experiments 1–3 indicate that the CNN network alone yields superior classification performance than other networks. Similarly, in experiments 4–6, performance noticeably declines when the CNN network is removed, underscoring the importance of integrating local and global features to extract diverse information from multimodal models.
	
	\item Experiments 2, 3, 5, and 6 demonstrate that a self-attention-based transformer alone is not effective, suggesting that relying solely on global feature extraction is limited. In contrast, the Mamba network outperforms the transformer, highlighting its advantage in modeling long sequences in spectral images.
	
	\item Comparisons between experiments 7–9 and the corresponding experiments 4–6 without diffusion feature guidance reveal that models incorporating additional sub-networks perform better, which confirms the significance of diverse features.
	
	\item Finally, experiments 10 and 11 show that removing the mutual learning module and the adaptive modality mask strategy leads to a decline in performance, thereby verifying the effectiveness of these modules. Experiment 12 further demonstrates that the proposed model achieves excellent results.
\end{enumerate}

\subsection{Feature Complementarity Visualization}
\label{visualization}

To verify that our classification architecture learns complementary representations, we visualize the per-branch features immediately before the final fusion layer. Fig. \ref{9vis} shows features extracted from the LCZ HK dataset (13 classes, C0–C12, 50 samples per class) projected onto two dimensions via t-distributed stochastic neighbor embedding (t-SNE). The t-SNE projection is generated from the first 50 principal components of the data, using a perplexity of 30 and 1000 iterations to ensure reproducible results.

The three branch features capture complementary structure. For example, CNN features show intermixing of classes C0 and C2 that are cleanly separated by Transformer features, while Mamba features produce distinct islands for classes such as C5 and C8 that appear split or overlapped in the other embeddings.
Notably, C11 is fragmented into multiple local modes in the CNN plot, whereas it becomes consolidated in the Transformer plot, and occupies largely non-overlapping regions in the Mamba plot. Similar complementary behaviours are observed for C3 and C7, suggesting that feature fusion may improve class separability.

\subsection{Classification Results}

To illustrate the effectiveness of the proposed BDGF, we conducted a comparative analysis with ten state-of-the-art multimodal classification models. AsyFFNet employs an asymmetric neural network with weight-sharing residual blocks for multimodal feature extraction and introduces a channel exchange mechanism with sparse constraints for feature fusion. CALC builds a multi-level feature fusion module and a spatial attention-guided discriminator based on CNNs and generative adversarial networks. SS-MAE adopts a similar network architecture but incorporates pre-training and masked self-supervised strategies. 

\begin{table*}
	\centering  
	\caption{Classification accuracy (\%) on the Augsburg dataset with 100 training samples for each class (bold values are the best and underline values are the second)} 
	\label{Table-a}
		\begin{tabular}{cccccccccccc} %第一列设置宽度为45pt 全为左对齐 没有分割线
			\toprule % 顶部线
			\toprule % 顶部线
			Class  & & AsyFFNet&CALC&Fusion-HCT  & MACN &NCGLF &UACL&SS-MAE& HLMamba&MSFMamba  & BDGF\\
			\midrule
			1 & &	94.32&	95.67&	96.32&	96.84&	92.32&	93.51&	\underline{97.52}&	97.51	&\textbf{97.78}	&97.03
			%			$\pm$0.80
			\\
			2 &	&85.80&	\underline{89.96}&	85.89&	86.89&	89.55&	89.94&	89.21&	89.07&	89.61&	\textbf{91.74}
			%			$\pm$\textbf{1.71}
			\\
			3 &	&\underline{85.28}&	51.37&	78.81	&73.21&	70.00&	64.09&	75.58&	35.54&	62.06&	\textbf{87.28}
			%			$\pm$\textbf{1.04}
			\\
			4 &	&91.51&	90.04&	94.32&	94.48&	94.38&	95.79&	95.56&	\textbf{96.25}&	94.73&	\underline{96.13}
			%			$\pm$\underline{0.82}
			\\
			5 & &96.84	&55.79	&96.84&	96.42&	93.96&	95.30&	\textbf{98.11}&	93.68&	96.63&	\underline{97.20}
			%			$\pm$\underline{0.86}
			\\
			6 & &	69.64&	\textbf{96.50}&	62.72&	61.55	&73.40&	\underline{90.03}&	71.52&	89.32&	66.60&	84.06
			%			$\pm$3.81
			\\
			7 & & 	\textbf{82.66}&	76.85&	74.20&	79.30&	80.03&	77.45&	80.98&	\underline{81.26}&	77.00&	77.41
			%			$\pm$1.64
			\\
			
			\midrule
			OA & &	88.91&	88.79&	89.65&	89.98&	90.23&	91.09&	\underline{91.73}&	90.31&	90.81&	\textbf{93.57}
			%			$\pm$\textbf{0.51}
			\\
			AA & &	86.58	&79.46&	84.15&	84.10&	84.81&	86.59&	\underline{86.93}&	83.23&	83.49&	\textbf{90.12}
			%			$\pm$\textbf{0.75}
			\\
			Kappa&& 	84.62&	84.29	&85.60&	86.01&	86.37&	87.53&	\underline{88.40}&	86.41&	87.13&	\textbf{90.93}
			%			$\pm$\textbf{0.70}
			\\
			\bottomrule % 底部线
			\bottomrule % 底部线
		\end{tabular}
\end{table*}

\begin{table*}
	\centering  % 显示位置为中间
	\caption{Classification accuracy (\%) on the Yellow River Estuary dataset with 100 training samples for each class (bold values are the best and underline values are the second)}  % 表格标题
	\label{Table-y}
		\begin{tabular}{cccccccccccc} %第一列设置宽度为45pt 全为左对齐 没有分割线
			\toprule % 顶部线
			\toprule % 顶部线
			Class  & & AsyFFNet&CALC&Fusion-HCT  & MACN &NCGLF &UACL&SS-MAE& HLMamba&MSFMamba  & BDGF\\
			\midrule
			1 &&	91.54&	86.14&	91.55&	\textbf{92.34}&	90.63&	\underline{91.92}&	90.48&	86.20&	88.74&	88.23
			%		$\pm$0.93
			
			\\
			2 & &	69.91&	\underline{73.48}&	67.55&	65.78&	71.03&	64.07&	69.11&	71.37&	72.28&	\textbf{76.72}
			%		$\pm$\textbf{2.50}
			
			\\
			3 &	&75.24&	64.48&	83.97&	86.25&	\textbf{88.98}&	71.67&	80.72&	88.84&	\underline{88.88}&	81.00
			%		$\pm$3.71
			
			\\
			4 & & 	\underline{79.78}&	71.87&	77.35&	79.65&	76.29&	77.00&	78.80&	\textbf{82.02}&	77.55&	75.57
			%		$\pm$1.59
			\\
			5 & &83.12&	\textbf{89.36}&	76.45&	74.01&	76.19&	80.85&	\underline{84.98}&	78.49&	80.28&	79.54\\
			%		$\pm$4.00
			
			\midrule
			OA& & 	76.45&	75.84&	75.70&	75.18&	77.97&	72.08&	76.81&	77.99&	\underline{78.78}	&\textbf{79.55}
			%		$\pm$\textbf{1.29}
			\\
			AA& &79.92	&77.07&	79.37&	79.60&	80.62&	76.90&	80.82&	\underline{81.38}&	\textbf{81.55}&	80.21
			%		$\pm$1.25
			\\
			Kappa&&	67.50&	66.17&	66.70&	66.08&	69.60&	62.16&	68.13&	69.63	&\underline{70.63}&	\textbf{71.00}
			%		$\pm$\textbf{1.65}
			\\
			\bottomrule % 底部线
			\bottomrule % 底部线
		\end{tabular}
\end{table*}

\begin{table*}
	\centering  % 显示位置为中间
	\caption{Classification accuracy (\%) on the LCZ HK dataset with 50 training samples for each class (bold values are the best and underline values are the second)}  % 表格标题
	\label{Table-hk}
		\begin{tabular}{cccccccccccc} %第一列设置宽度为45pt 全为左对齐 没有分割线
			\toprule % 顶部线
			\toprule % 顶部线
			Class  &  & AsyFFNet&CALC&Fusion-HCT  & MACN &NCGLF &UACL&SS-MAE& HLMamba&MSFMamba  & BDGF\\
			\midrule
			1 & & 64.37&	77.62&	72.46&	53.53	&76.40&	68.67&	78.14&	74.87&	\textbf{83.30}&	\underline{80.77}
			%			$\pm$\underline{3.20}
			
			\\
			2 & &	95.35&	66.67&	80.62&	86.82	&90.67&	34.11&	89.15&	\underline{96.12}&	73.64&	\textbf{97.05}
			%			$\pm$\textbf{2.05}
			
			\\
			3 & & 89.13&	97.10&	\underline{97.46}&	97.46&	93.83&	97.10&	92.03&	93.48&	96.38&	\textbf{99.02}
			%			$\pm$\textbf{0.67}
			
			\\
			4 & & 	80.90&\underline{93.74}&	81.70&	91.97&	87.31&	78.97&	\textbf{93.94}&	76.08	&86.84&	90.16
			%			$\pm$1.97
			
			\\
			5 & & 97.37	&94.74&	\underline{98.68}&	97.37&	\underline{99.68}&	98.68&	94.74&	89.47&	93.42&	\textbf{99.87}
			%			$\pm$\textbf{0.39}
			
			\\
			6 & & 	\textbf{100.00}&	\textbf{100.00}&	92.86&	98.57&	\underline{99.33}&	87.14&	98.57&	98.57&	88.57&	97.14
			%			$\pm$0.14
			
			\\
			7 & & 95.40&	93.10&	94.25&	94.25&	95.84&	95.40&	\textbf{100.00}&	83.06&	96.55&\underline{99.66}
			%			$\pm$\underline{0.53}
			
			\\
			8 & & 93.49&	96.45&	89.94&	98.22&	96.67&	\textbf{100.00}&	97.63&	94.08&	92.31&	\underline{98.58}
			%			$\pm$\underline{0.71}
			
			\\
			9 & &94.70&	\textbf{98.08}&	94.76&	94.76&	97.17&	96.87&	95.59&	93.04&	94.89&	\underline{97.48}
			%			$\pm$\underline{2.04}
			
			\\
			10 & &	82.45&	87.76&	84.90&	84.90&	87.02	&84.90&	\underline{90.82}&	80.41&	61.02&	\textbf{91.94}
			%			$\pm$\textbf{1.55}
			
			\\
			11 & & 71.76&	91.11&	92.82&	82.37&	\underline{94.63}&	80.97&	90.02&	87.99&	86.90&	\textbf{96.93}
			%			$\pm$\textbf{0.80}
			
			\\
			12 & & 81.28&	77.22&	85.78&	74.65&	85.01&	\underline{89.73}&	84.17&	68.66&	87.38&	\textbf{93.52}
			%			$\pm$\textbf{1.24}
			
			\\
			13 & & 	98.12&	97.61&	96.91&	\textbf{99.69}&	97.07&	\underline{99.61}&	99.10&	98.55&	97.57&	98.53
			%			$\pm$0.69
			
			\\
			\midrule
			OA& & 88.38&	91.98&	90.87&	89.40&	92.51	&90.59&	\underline{93.16}&	88.26&	90.40&	\textbf{95.35}
			%			$\pm$\textbf{0.58}
			
			\\
			AA& &	88.03&	90.09&	89.47&	88.81&	92.36&	88.73&	\underline{92.59}&	87.41&	87.60&	\textbf{95.43}
			%			$\pm$\textbf{0.37}
			
			\\
			Kappa&&	86.09	&90.38&	89.09&	87.29	&91.16&	85.55&	\underline{91.79}&	85.95&	88.49&	\textbf{94.42}
			%			$\pm$\textbf{0.69}

			\\
			\bottomrule % 底部线
			\bottomrule % 底部线
			
		\end{tabular}
\end{table*}
\begin{table*}
	\centering  % 显示位置为中间
	\caption{Classification accuracy (\%) on the Berlin dataset with 100 training samples for each class (bold values are the best and underline values are the second)}  % 表格标题
	\label{Table-b}
	\begin{tabular}{cccccccccccc} %第一列设置宽度为45pt 
		%全为左对齐 没有分割线
		\toprule
		\toprule % 顶部线
		Class  & &  AsyFFNet&CALC&Fusion-HCT  & MACN &NCGLF &UACL&SS-MAE& HLMamba&MSFMamba  & BDGF\\
		\midrule % 中部线
		1 &&	81.29&	81.65&	79.67&	84.96&	\underline{89.86}&	88.93&	88.14&	89.66&	\pmb{89.87}&	82.34
		%			$\pm$5.48
		\\
		2 &&	69.62&	\underline{71.16}&	68.04&	65.84&	67.63	&\pmb{72.06}	&67.46&	66.93&	68.73&	70.87
		%			$\pm$3.81
		\\
		3 && 62.05&	64.68&	64.73&	64.94&	67.65&	66.99	&\underline{68.08}	&67.13&	67.52&	\pmb{76.81}
		%			$\pm$\pmb{3.95}
		\\
		4&&	82.04&	84.25&	\pmb{89.51}&	86.72&	85.69&	82.55&	\underline{87.47}&	80.80&	84.04	&85.97
		%			$\pm$3.13
		\\
		5&&93.06&	94.01&	96.09&	94.37&	95.27&	88.96&	93.04&	92.81&	\pmb{96.66}&	\underline{96.12}
		%			$\pm$\underline{2.35}
		\\
		
		6&&	74.09&	84.97&	80.33	&\underline{85.02}	&\pmb{86.69}	&53.00&	83.88&	80.83&	83.18&	74.95
		%			$\pm$7.08
		\\
		
		7&&	60.12&	51.31&	57.20&	52.93&	\pmb{62.16}&	21.66&	\underline{60.98}&	54.02&	60.04&	57.95
		%			$\pm$4.17
		\\
		
		8&&	91.19&	92.82&	92.44&	80.14&	\underline{94.79}&	\pmb{95.05}&	93.39&	92.79&	85.48&	94.49
		%			$\pm$2.02
		\\
		
		\midrule
		OA  & & 	73.07&	74.29&	73.18&	71.85&	74.24&	72.90&	73.93&	72.44&	\underline{74.36}&	\pmb{75.11}
		%			$\pm$\pmb{1.49}
		
		\\
		AA  & &	76.68&	78.11&	78.50&	76.86&	\pmb{81.22}&	61.26&	80.31&	78.12&	79.44&	\underline{79.94}
		%			$\pm$\underline{1.90}
		
		\\
		Kappa&&	61.83&	63.72&	62.64&	61.16&	64.55	&\pmb{71.15}&	63.76&	61.84&	64.29&	\underline{64.73}
		%			$\pm$\underline{1.58}
		
		\\
		\bottomrule % 底部线
		\bottomrule
		%\hline  % 表格的横线
	\end{tabular}
\end{table*}

\begin{figure}[htp]
	\includegraphics[width=\linewidth]{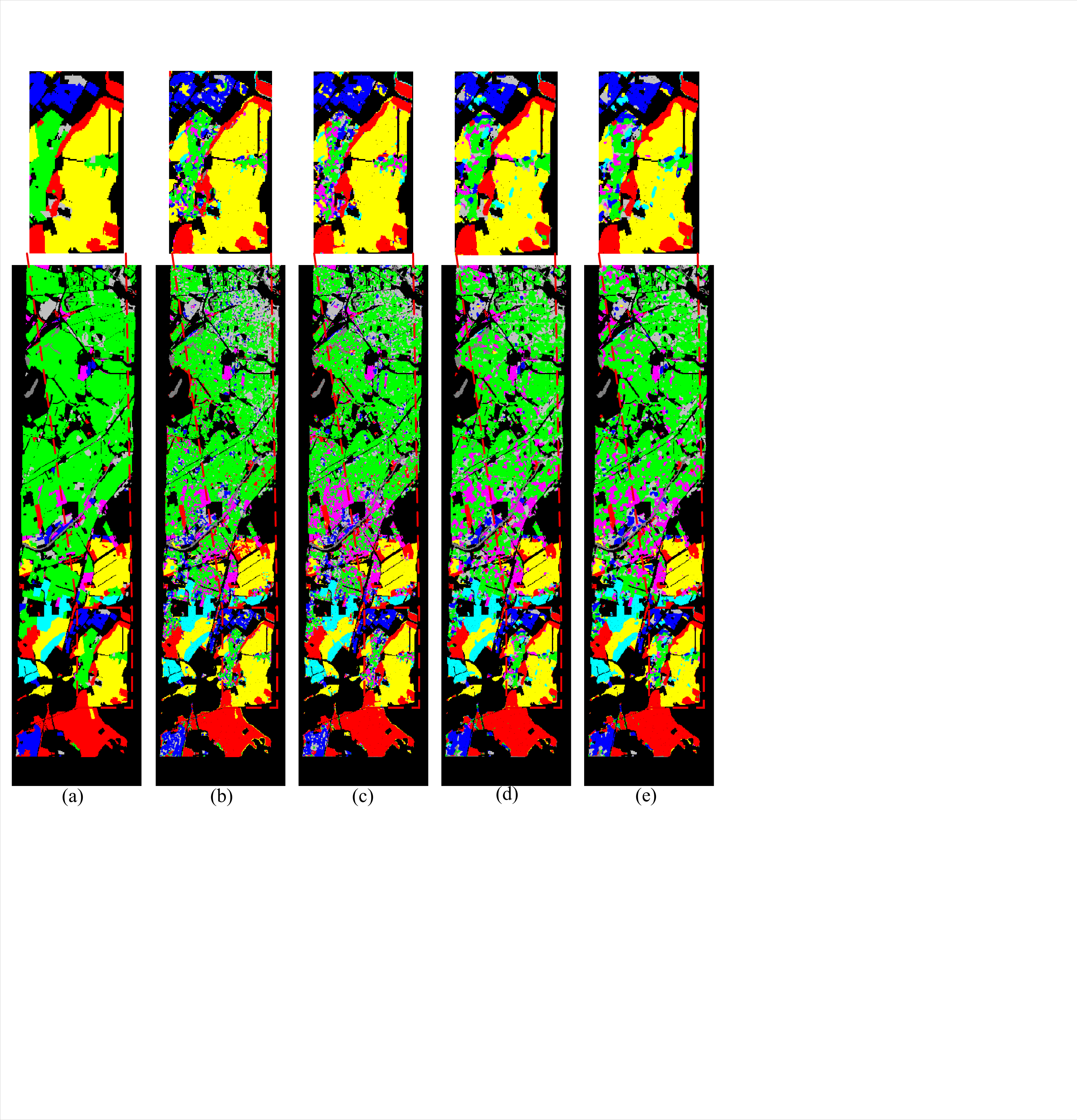}
	
	\caption{Classification maps and OA\% obtained on the Berlin dataset using several methods.  (a) Ground-truth map. (b) SS-MAE (73.93\%). (c) HLMamba (72.44\%). (d) MSFMamba (74.36\%). (e) BDGF (75.11\%).}
	\label{fig10}
\end{figure}

\begin{figure}[htp]
	\includegraphics[width=\linewidth]{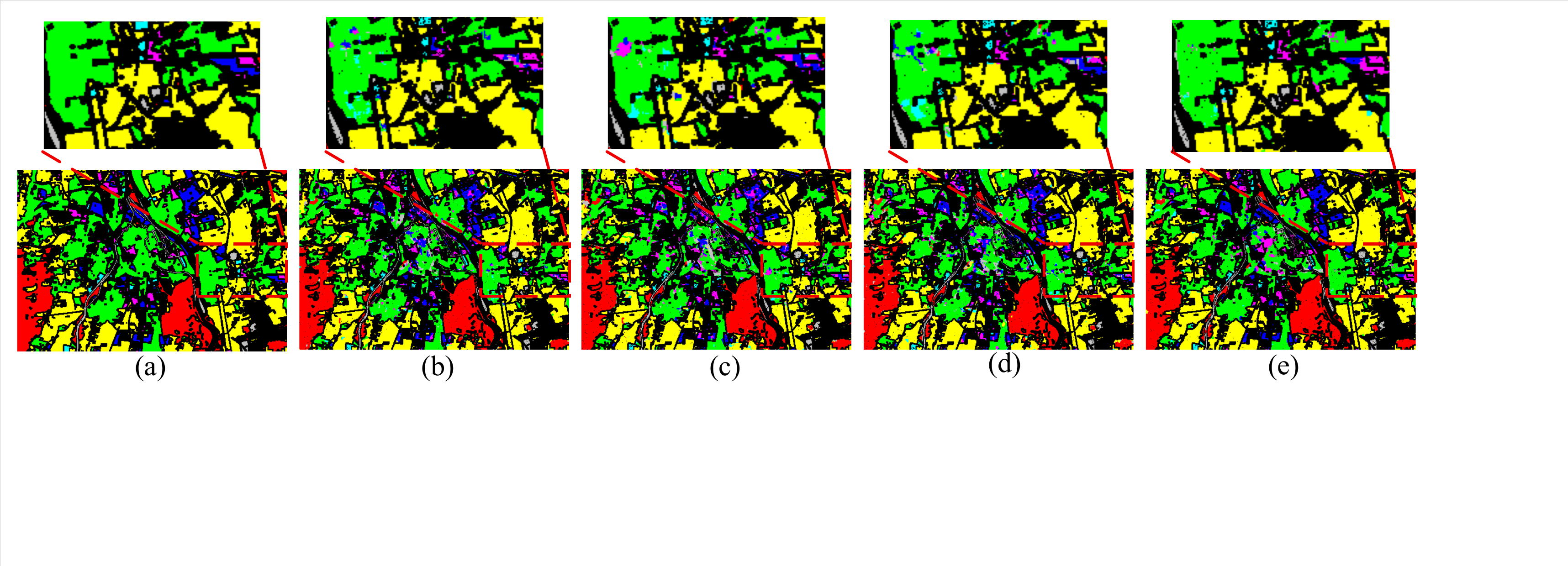}
	\centering
	\caption{Classification maps and OA\% obtained on the Augsburg dataset using several methods. (a) Ground-truth map. (b) SS-MAE (91.73\%). (c) HLMamba (90.31\%). (d) MSFMamba (90.81\%). (e) BDGF (93.57\%).}
	\label{fig9}  
\end{figure}

\begin{figure}[htp]
	\includegraphics[width=\linewidth]{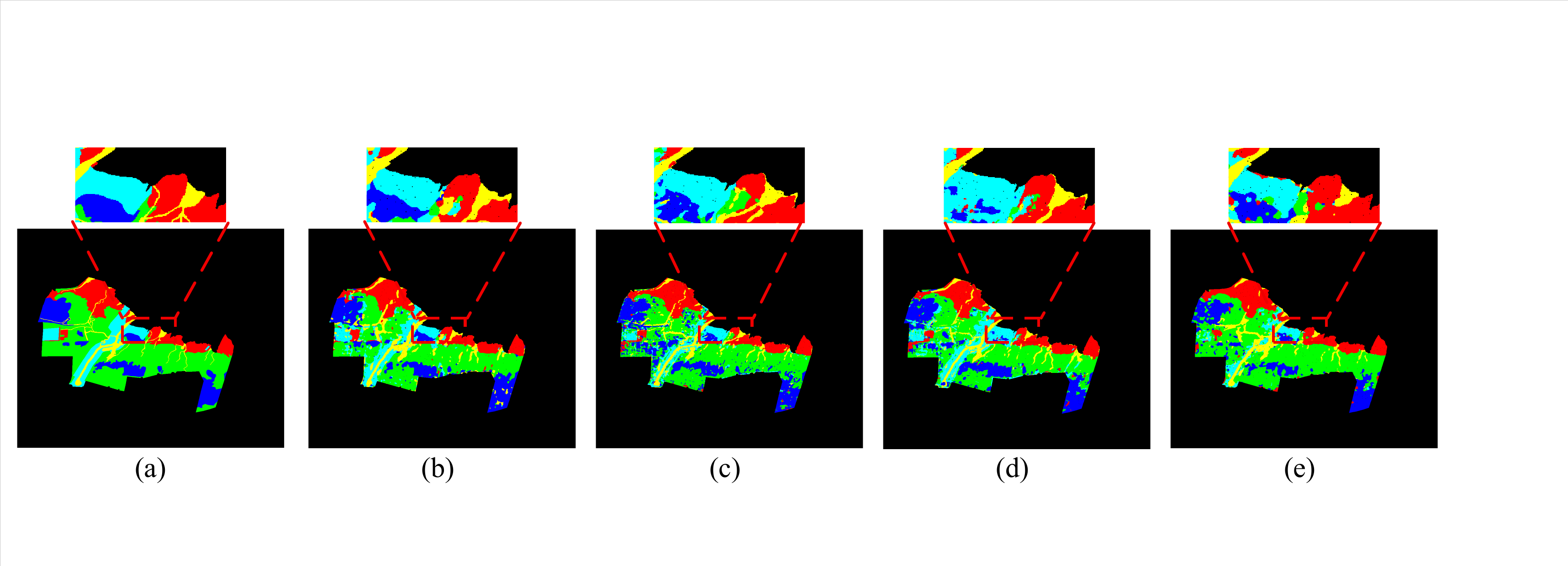}
	\centering
	\caption{Classification maps and OA\% obtained on the Yellow River Estuary dataset using several methods. (a) Ground-truth map. (b) SS-MAE (76.81\%). (c) HLMamba (77.99\%) (d) MSFMamba (78.78\%). (e) BDGF (79.55\%).}
	\label{fig11}
\end{figure}

\begin{figure}[htp]
	\includegraphics[width=\linewidth]{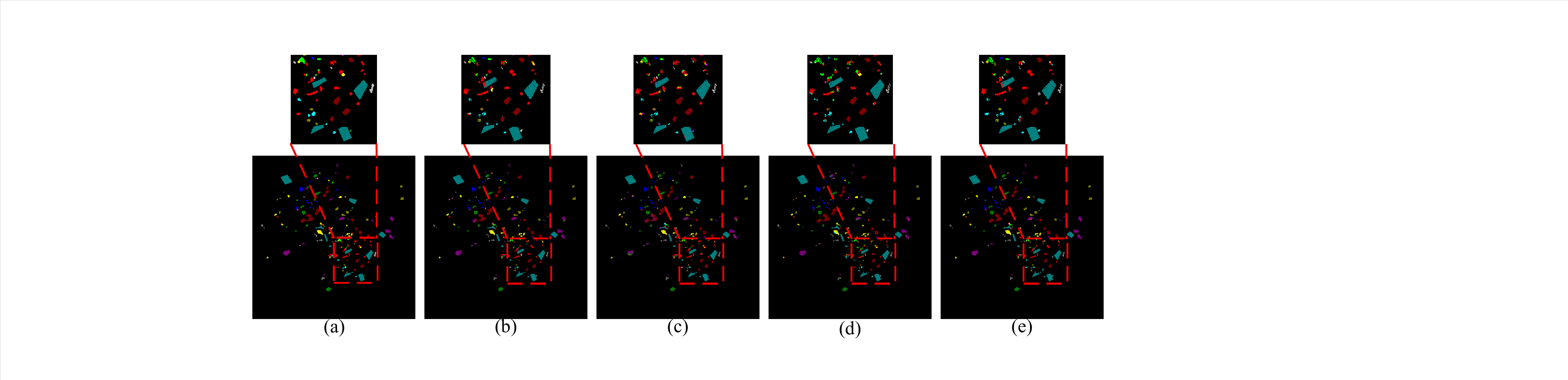}
	\centering
	\caption{Classification maps and OA\% obtained on the LCZ HK dataset using several methods. (a) Ground-truth map. (b) SS-MAE (93.16\%). (c) HLMamba (88.26\%) (d) MSFMamba (90.40\%). (e) BDGF (95.35\%).}
	\label{fig12} 
\end{figure}

Furthermore, we selected four methods that focus on multi-branch network structures. Fusion-HCT and MACN integrate CNNs and transformers to capture both local and global features, introducing innovative attention mechanisms for multimodal feature fusion. NCGLF enhances CNN and transformer structures with structural information learning and invertible neural networks. UACL proposes a contrastive learning strategy to access reliable multimodal samples. In addition, two recent multimodal learning methods based on the multi-scale Mamba structure are included for comparison. HLMamba constructs a multimodal Mamba fusion module and introduces a gradient joint algorithm to enhance modality information, while MSFMamba employs spatial, spectral, and fused Mamba branches with a large effective receptive field to achieve multi-scale feature fusion.

The performance of these classification methods is summarized in Tables \ref{Table-a}--\ref{Table-b}. For a visual comparison, the respective segmentation maps of several methods are presented in Figs. \ref{fig10}-\ref{fig12}. Based on these outcomes, the following conclusions can be drawn:
\begin{enumerate}
	
	\item Methods that integrate multi-scale and multi-branch architectures for multimodal data fusion exhibit superior classification performance. Among these, NCGLF outperforms the two CNN-based methods due to its effective integration of global and local information.
	\item SS-MAE, which learns multimodal features through a reconstructed pre-training paradigm, achieved suboptimal classification results on the Augsburg and LCZ HK datasets. Similarly, MSFMamba, which employs a multi-scale network with Mamba as its core, obtained suboptimal results on the other two datasets.
	\item Leveraging guidance from robust diffusion features, the proposed BDGF improves collaborative learning of multimodal features across different branches, as evidenced by its highest OA index along with excellent AA and kappa values. On the four datasets, BDGF consistently outperforms previous state-of-the-art models by 1.84\%, 0.77\%, 2.19\%, and 0.75\% in OA index, respectively. 
	
%	As one can see from the figures and the tables, in the Augsburg and Berlin datasets, the proposed BDGF achieves high classification results in the “residential area” and “industrial area” categories. Among the 13 categories in the LCZ HK dataset, the BDGF has the highest semantic classification accuracy on 6 classes.
\end{enumerate}

\subsection{Uncertainty Analysis}

\begin{table*}[htp]
	\centering
	\caption{Mean, stand deviation, CV, 95\% CI, and NSI of the proposed BDGF across 10 runs on the four considered datasets.}
	\resizebox{\textwidth}{!}{
		\begin{tabular}{c|ccc|ccc|ccc|ccc}
			\toprule
			Metric & \multicolumn{3}{c}{LCZ HK} & \multicolumn{3}{c}{Yellow River} & \multicolumn{3}{c}{Berlin} & \multicolumn{3}{c}{Augsburg} \\
			\cline{2-13}
			& OA & AA & Kappa & OA & AA & Kappa & OA & AA & Kappa & OA & AA & Kappa \\
			\hline
			Mean (\%) & 95.35 & 95.43 & 94.42 & 79.55 & 80.21 & 71.00 & 75.11 & 79.94 & 64.73 & 93.57 & 90.12 & 90.93 \\
			Std (\%) & 0.61 & 0.39 & 0.73 & 1.36 & 1.32 & 1.74 & 1.57 & 2.01 & 1.66 & 0.54 & 0.79 & 0.74 \\
			CV (\%) & 0.64 & 0.41 & 0.77 & 1.71 & 1.65 & 2.46 & 2.09 & 2.51 & 2.57 & 0.58 & 0.88 & 0.81 \\
			95\%CI (\%) & $\pm$0.44 & $\pm$0.28 & $\pm$0.52 & $\pm$0.97 & $\pm$0.95 & $\pm$1.25 & $\pm$1.13 & $\pm$1.44 & $\pm$1.19 & $\pm$0.39 & $\pm$0.57 & $\pm$0.53 \\
			NSI (\%) & 0.0064 & 0.0041 & 0.0077 & 0.0171 & 0.0165 & 0.0246 & 0.0209 & 0.0251 & 0.0257 & 0.0058 & 0.0088 & 0.0081 \\
			\bottomrule
	\end{tabular}}
	\label{uncertain_table}
\end{table*}

\begin{figure}[t]
	\includegraphics[width=\linewidth]{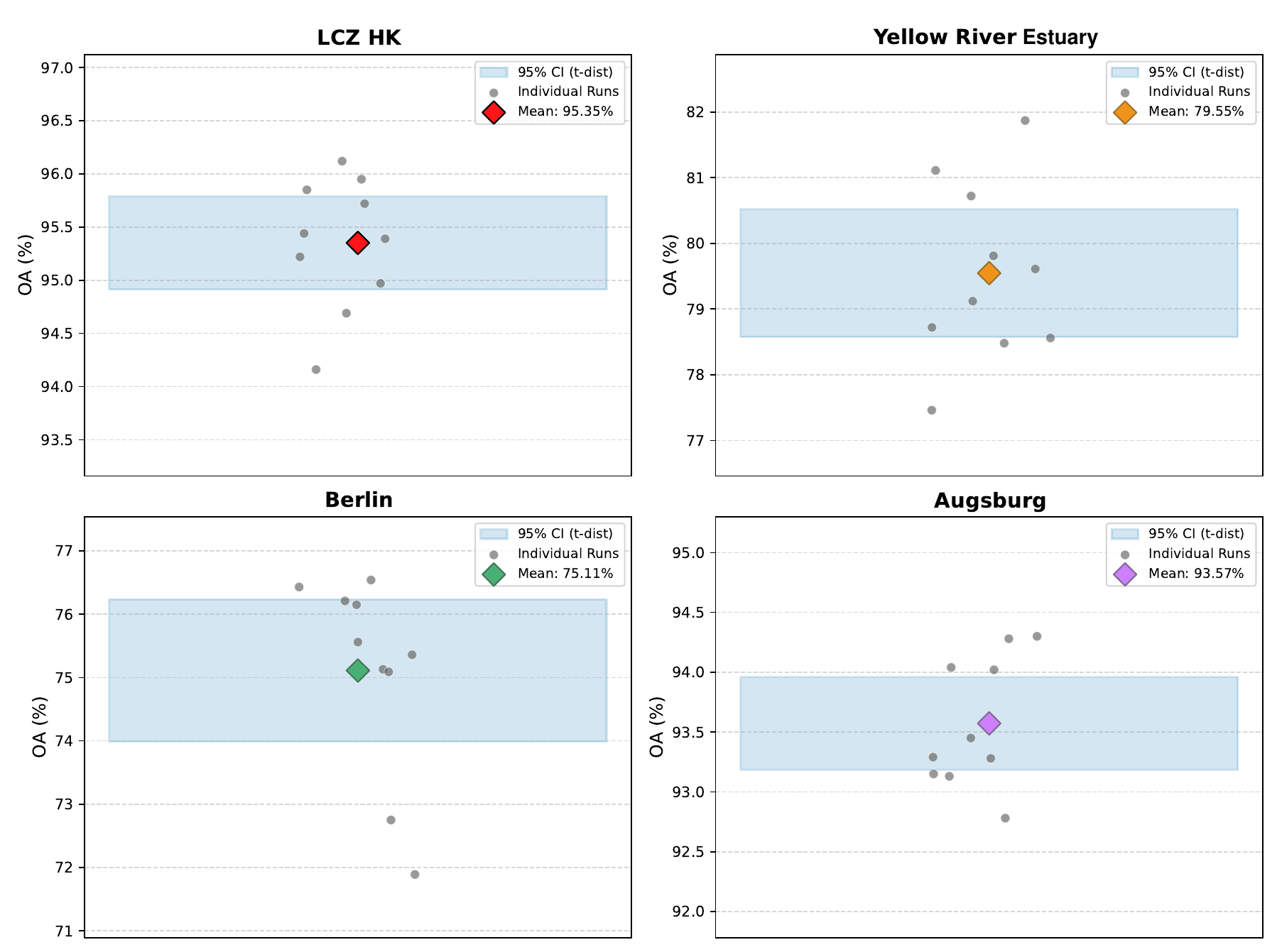}
	\centering
	\caption{Individual values, mean, and 95\% CI of OA (\%) across 10 runs on the four considered datasets.} 
	\label{uncertainty_figure}
\end{figure}
\begin{figure}
	\includegraphics[width=\linewidth]{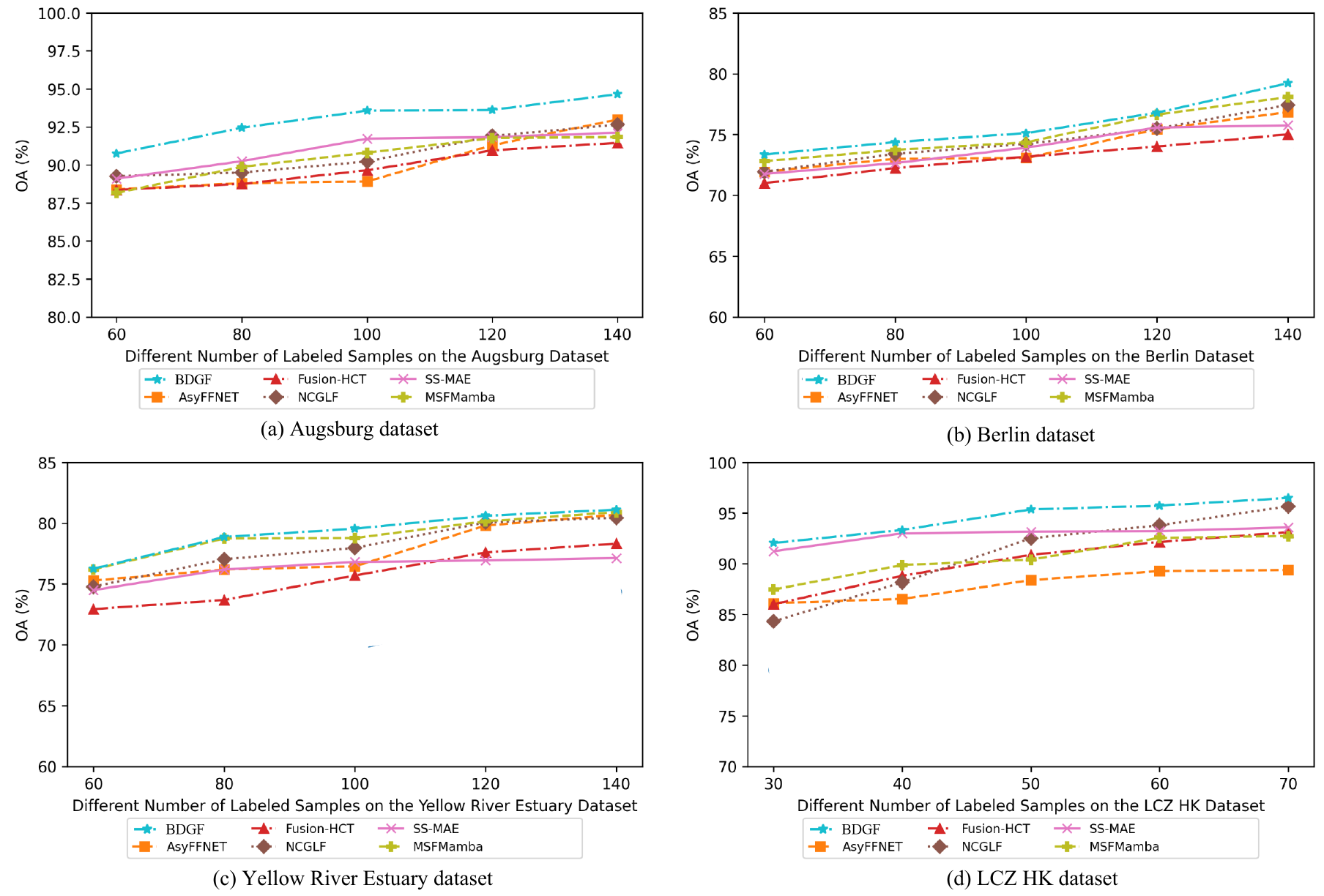}
	\caption{OA\% versus the number of labeled samples on the four considered datasets. (a) Augsburg dataset. (b) Berlin dataset. (c) Yellow River Estuary dataset. (d) LCZ HK dataset.}
	\label{fig13}
\end{figure}

To assess the statistical reliability and generalization capability of the proposed BDGF, we conduct an extensive uncertainty analysis spanning 10 independent experimental runs for each dataset, maintaining random seeds 0-9. As summarized in Table~\ref{uncertain_table}, we evaluate model uncertainty using standard deviation, Coefficient of Variation (CV), Normalized Sensitivity Index (NSI), and the 95\% Confidence Interval (CI) calculated via the t-distribution. Fig.~\ref{uncertainty_figure} shows the individual values, mean, and confidence intervals of the OA across 10 runs. BDGF consistently demonstrated very low variance, with the CV remaining below 5\% across all multimodal datasets.

To further verify the effectiveness of the model on different training samples, we conducted experiments with 60, 80, 120, and 140 labeled samples per class for HSI and SAR classification, and 30, 40, 60, and 70 labeled samples per class for MSI and SAR classification. The performance of each method under these settings is illustrated in Fig. \ref{fig13}. Different methods exhibited differing levels of sensitivity to the number of labeled samples. However, across all configurations, the proposed BDGF framework consistently achieved the highest classification accuracy.

\subsection{Transferability Analysis}
\begin{table}[ht]
	\centering
	\caption{Classification accuracy (\%) in the cross-model feature transfer experiments on the four considered datasets}
	\label{transfer}
	\footnotesize
	\resizebox{0.48\textwidth}{!}{
		\begin{tabular}{@{}lcccccc@{}}
			\toprule
			\toprule
			& \multicolumn{6}{c}{\textbf{Attention Classified Module}} \\
			\cmidrule(lr){2-7}
			& \multicolumn{3}{c}{Masking Diff} & \multicolumn{3}{c}{Concatenate Diff} \\
			\cmidrule(lr){2-4} \cmidrule(lr){5-7}
			Dataset & OA & AA & Kappa & OA & AA & Kappa \\
			\midrule
			Augsburg & 70.72 & 56.41 & 46.34 & 67.15 & 64.06 & 55.64 \\
			Berlin & 63.73 & 65.05 & 54.56 & 42.24 & 22.44 & 18.35 \\
			Yellow River Estuary & 68.26 & 64.78 & 59.49 & 63.04 & 48.23 & 38.74 \\
			LCZ HK & 91.75 & 93.12 & 90.12 & 91.24 & 91.40 & 89.71 \\
			\midrule[0.5pt]
			& \multicolumn{6}{c}{\textbf{Group Network}}\\
			\cmidrule(lr){2-7}
			& \multicolumn{3}{c}{Masking Diff} & \multicolumn{3}{c}{Concatenate Diff} \\
			\cmidrule(lr){2-4} \cmidrule(lr){5-7}
			Dataset & OA & AA & Kappa & OA & AA & Kappa \\
			\midrule
			Augsburg & 93.57 & 90.12 & 90.93 & 92.62 & 89.90 & 89.59 \\
			Berlin  & 75.11 & 79.94 & 64.73 & 74.30 & 77.38 & 63.50 \\
			Yellow River Estuary& 79.55 & 80.21 & 71.00 & 77.85 & 80.16 & 69.02 \\
			LCZ HK & 95.35 & 95.43 & 94.42 & 95.06 & 95.07 & 94.07 \\
			\bottomrule
			\bottomrule
	\end{tabular}}
\end{table}

\begin{table*}[t]
	\centering
	\caption{Number of parameters (M, Million) and GFLOPS of different considered methods}
	\label{TIME}
	\resizebox{\textwidth}{!}{
		\begin{tabular}{cccccccccccccc}
			\toprule
			\toprule
			& &  AsyFFNET&CALC&Fusion-HCT&MACN&NCGLF&UACL&HLMamba&	MSFMamba&SS-MAE (P)&SS-MAE (T)	&BDGF (P)&	BDGF (T)\\
			\hline
			\multirow{2}*{Augsburg}
			& Params. (M) &1.08	&0.94&	0.43&	0.17&	0.44&	0.19&	0.23&	0.82	&	7.72&	4.51&	4.30&	16.38
			\\		
			& GFLOPs  & 17.76&	7.23&	0.59&	0.70&	8.72&	2.38&	22.68&	25.17	&	74.21&	51.34&	996.43&	99.33
			\\	
			\hline
			\multirow{2}*{Yellow River Estuary}
			%\cline{2-4}
			& Params. (M) &1.08	&0.92&	0.43&	0.17&	0.44&	0.18&	0.20&	0.78&		7.70&	4.50&	4.30&	9.56\\
			& GFLOPs  &17.72&	6.80&	0.59&	0.70&	8.72&	2.24&	14.67&	25.15&		69.54&	48.13&	498.22&	64.28\\
			\hline
			\multirow{2}*{Berlin}
			& Params. (M) &1.06&	0.99&	0.43&	0.17&	0.44&	0.21&	0.19&	2.46	&	7.80&4.54&	4.30&	16.38
			\\
			& GFLOPs  &17.35&	9.04&	0.59&	0.70&	8.72&	2.99&12.77	&61.91&		94.69&	65.02&	996.43&	99.33
			\\
			\hline
			\multirow{2}*{LCZ HK}
			%\cline{2-4}
			& Params. (M) &1.06	&0.79&	0.43&	0.07&	0.34&	0.13&	0.18&	0.21	&	7.56&	4.44&	4.30&	6.15
			\\
			& GFLOPs  &17.32&	2.47&	0.59&	0.37&	7.07&	0.80&11.58&	3.83&		23.29&	17.15&	124.55&	46.12
			\\		
			\bottomrule
			\bottomrule
	\end{tabular}}
\end{table*}

Different from the fine-tuning training paradigm such as SS-MAE, the proposed BDGF framework employs an unsupervised diffusion process to learn the joint data distribution of multimodal images and directly leverages these learned features for downstream classification. To evaluate both the transferability of our adaptive masking strategy and the modularity of our group network, we perform cross-model feature-transfer experiments with SpectralDiff \cite{10234379}, which shares a similar training paradigm. Specifically, we first adapt SpectralDiff to multimodal inputs using early fusion via channel-wise concatenation \cite{9174822,tu2024ncglf2,li2022deep}. From the pre-trained diffusion backbones of both methods, we extract features and graft them into the other model’s classification head. In Table \ref{transfer}, the pre-trained diffusion model and sub-network of SpectralDiff are denoted as “Concatenate Diff” and “Attention Classified Module”, respectively, while the corresponding annotations of our method are “Masking Diff” and “Group Network”.

As one can see from the Table \ref{transfer}, with the same classifier backbone, the adaptive masking strategy consistently enhances performance. Moreover, using identical pre-trained diffusion features, the group network outperforms SpectralDiff’s attention-based classification module across all datasets. Notably, on the Augsburg dataset, the attention classification module suffers a substantial accuracy drop, which shows the superior robustness of our group fusion in capturing diverse features.

\subsection{Computational Complexity Analysis}

In this section, we evaluate the computational complexity of all models in terms of GFLOPs and number of parameters (millions). Table \ref{TIME} presents these metrics for the four datasets considered. Note that SS-MAE and BDGF employ a pre-training paradigm, and (P) and (T) denote pre-training and training phases, respectively. Under unsupervised learning, larger models generally generalize better. Pre-training methods incur higher computational costs than other approaches, and adopting a dual-branch structure for the pre-training autoencoder and diffusion model further increases complexity. SS-MAE and BDGF mitigate this issue by merging multimodal images along the channel dimension. Their superior classification results across the four datasets confirm the effectiveness of pre-training. For the proposed BDGF, the parameter count arises from integrating multi-branch sub-networks, while the FLOPs reflect the diffusion model’s pre-training. Although BDGF exhibits higher computational complexity, it remains within acceptable limits and achieves the best classification performance.

\section{Conclusion}
In this paper, we have proposed BDGF framework for multimodal remote sensing image classification. BDGF leverages robust diffusion features to guide a group network that integrates local, global, and sequential features. An adaptive modality masking strategy is introduced to mitigate modality imbalance during pre-training, ensuring a balanced representation between spectral and SAR images. In addition, the diffusion features are hierarchically fused through feature fusion, group attention mechanism, and cross-attention mechanism. Finally, a mutual learning strategy coordinates the predictions of each sub-network to improve the overall performance.

Extensive experiments on four multimodal remote sensing datasets validate the effectiveness of BDGF. Ablation studies confirm the contribution of each feature guidance module and strategy, and comparative evaluations under varying numbers of labeled samples demonstrate that BDGF outperforms baseline methods in terms of classification accuracy. In addition,  in cross-model feature-transfer experiments with SpectralDiff, the two-stage BDGF exhibits robust transferability. Furthermore, visualizations of diversity features on the LCZ HK dataset demonstrate the complementary nature of the proposed method’s features. Computational complexity analysis shows that pre-training models are costly, underscoring the efficiency gains of a single-branch pre-training strategy.

While BDGF improves multimodal feature fusion and classification accuracy, there remains room for further enhancement. 
Future work will focus on more efficient pre-training paradigms and multi-task learning methods to enhance inter-network collaboration. 
In addition, the framework will be further extended to other remote sensing applications, such as scene classification and change detection.
%\printcredits
	\section*{Data availability}
The code and data used in this study are available at \url{https://github.com/HaoLiu-XDU/BDGF}.

%	\section*{Declaration of competing interest}
%	The authors declare that they have no known competing financial interests or personal relationships that could have appeared to influence the work.
\section*{Acknowledgements}
This work was supported by the China Scholarship Council (Grant No. 202406960026).

\section*{Declaration of generative AI and AI-assisted technologies in the writing process}
During the preparation of this work the authors used ChatGPT in order to polish the language. After using this tool, the authors reviewed and edited the content as needed and take full responsibility for the content of the publication article.
%% Loading bibliography style file
%\bibliographystyle{model1-num-names}
%\bibliographystyle{cas-model2-names}
\bibliographystyle{elsarticle-num}
% Loading bibliography database
\bibliography{cas-refs}

%\vskip3pt

\end{document}